\documentclass{article} 
\usepackage{iclr2027_conference,times}

\usepackage{amsmath,amsfonts,bm}

\def\eqref#1{equation~\ref{#1}}

\def\1{\bm{1}}

\DeclareMathAlphabet{\mathsfit}{\encodingdefault}{\sfdefault}{m}{sl}
\SetMathAlphabet{\mathsfit}{bold}{\encodingdefault}{\sfdefault}{bx}{n}

\usepackage{booktabs}
\usepackage{multirow}
\usepackage{array}
\usepackage{makecell}
\usepackage[table]{xcolor}
\usepackage{graphicx}
\usepackage{wrapfig}
\usepackage{tikz}
\usetikzlibrary{calc}

\definecolor{rankfirst}{HTML}{C7C7FF}   
\definecolor{ranksecond}{HTML}{E8E8FF}  

\newcommand{\first}[1]{\cellcolor{rankfirst}#1}
\newcommand{\second}[1]{\cellcolor{ranksecond}#1}
\newcommand{\third}[1]{#1}              

\usepackage{capt-of}
\usepackage{float}
\usepackage{adjustbox}
\usepackage{booktabs}
\usepackage{multirow}
\usepackage{placeins}

\newcolumntype{L}[1]{>{\raggedright\arraybackslash}m{#1}}
\newcolumntype{C}[1]{>{\centering\arraybackslash}m{#1}}

\usepackage{floatflt}
\usepackage{wrapfig}
\usepackage{adjustbox}
\usepackage{url}
\usepackage{amssymb}
\usepackage{xcolor}
\usepackage{graphicx}
\usepackage{xspace}
\usepackage{caption}
\definecolor{customcitecolor}{HTML}{3E7FA3}
\definecolor{customurlcolor}{HTML}{5878BE}
\definecolor{urlcolor}{RGB}{237, 2, 140}
\definecolor{brightred}{RGB}{255, 0, 0}

\definecolor{MethodColor}{HTML}{2F7F9D}

\newcommand{\method}{%
    \textcolor{MethodColor}{%
        \textsc{Vggt}\textit{-Diff}%
    }\xspace
}

\usepackage[colorlinks=true,
            citecolor=customcitecolor, 
            linkcolor=brightred, 
            urlcolor=urlcolor,
            backref=page]{hyperref}

\title{Visual Geometry Meets Diffusion for Sparse-View Novel View Synthesis}

\author{{\fontsize{8.8pt}{9.4pt}\selectfont\bfseries
Kangjie Chen\textsuperscript{1}\thanks{Equal contribution.}\hspace{0.7em}
Xiangyu Li\textsuperscript{1}\footnotemark[1]\hspace{0.7em}
Dongbin Zhang\textsuperscript{1}\hspace{0.7em}
Chaoda Zheng\textsuperscript{1}\hspace{0.7em}
Shijia Chen\textsuperscript{1}\hspace{0.7em}
Jinhao Deng\textsuperscript{1}
}\\[-0.05em]
{\fontsize{8.8pt}{9.4pt}\selectfont\bfseries
Hongbin Lin\textsuperscript{2}\hspace{0.7em}
Choo Sin Wai\textsuperscript{3}\hspace{0.7em}
Minqi Wang\textsuperscript{2}\hspace{0.7em}
Minghao Yang\textsuperscript{3}\hspace{0.7em}
Dake Zhong\textsuperscript{3}\hspace{0.7em}
Guorui Song\textsuperscript{3}
}\\[-0.05em]
{\fontsize{8.8pt}{9.4pt}\selectfont\bfseries
Yu Zhang\textsuperscript{1}\hspace{0.7em}
Xianming Liu\textsuperscript{1}\hspace{0.7em}
Boyang Wang\textsuperscript{1}\thanks{Corresponding author.}
}\\[0.20em]
{\fontsize{8.6pt}{9.2pt}\selectfont\normalfont
\textsuperscript{1}XPeng Motors
\hspace{1.2em}
\textsuperscript{2}The Chinese University of Hong Kong
\hspace{1.2em}
\textsuperscript{3}Tsinghua University
}
}

\iclrfinalcopy 
\begin{document}
\maketitle
\fancyhead[L]{Preprint}


\noindent
\begin{minipage}{\textwidth}
\centering

\begin{tikzpicture}[inner sep=0, outer sep=0]

\node[anchor=north west] (qualitative) at (0,0) {
    \includegraphics[
        width=0.55\textwidth
    ]{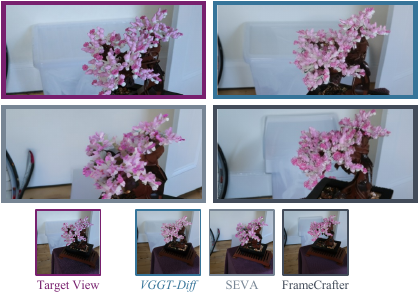}
};

\node[anchor=north west] (radar) at
    ([xshift=0\textwidth]qualitative.north east) {
    \includegraphics[
        width=0.402\textwidth
    ]{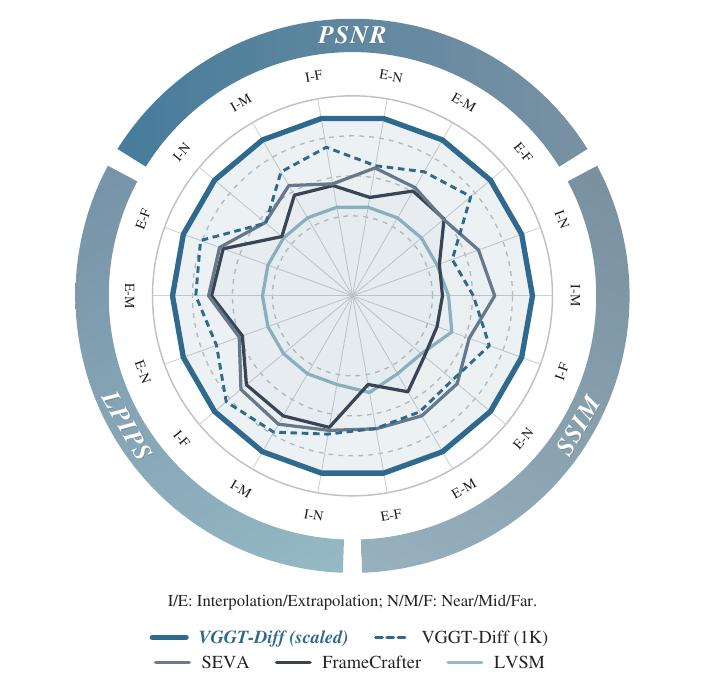}
};

\end{tikzpicture}

\vspace{-0.7em}

\captionof{figure}{
\textbf{\method{} grounds a pretrained video diffusion prior with
geometry-routed visual features for sparse-view novel-view synthesis.}
It maintains strong target-view fidelity across varying pose difficulties
while better preserving source-observed details than diffusion-based
baselines.
}
\label{fig:teaser}

\end{minipage}

\begin{abstract}
We present \method, a geometry-routed multi-view diffusion model for sparse-view novel view synthesis. Existing novel view synthesis (NVS) methods face a fundamental trade-off: reconstruction-based approaches preserve observed geometry but struggle to synthesize unseen regions, while diffusion-based methods provide strong generative priors yet rely on implicit source-to-query correspondence. \method{} bridges these regimes by routing visual geometry latents from VGGT-$\Omega$ into a pretrained video diffusion model. Each visual token is associated with a 3D point and confidence, then
transformed into query-aligned latent conditions through a confidence-aware Visual Geometry Router (VGR) that preserves front and back surface evidence. These conditions guide joint target-view denoising, while Point-Track Residual Consistency (PTRC) regularizes predicted-clean residuals along
reliable 3D tracks, improving multi-view stability. We further introduce robust geometry conditioning, combining training-time regularization with inference-time guidance for improved
robustness. Experiments show competitive or state-of-the-art performance across interpolation and extrapolation under different viewpoint difficulties. Our code is available at
\url{https://github.com/chenkangjie1123/VGGT-Diff}.
\end{abstract}

\section{Introduction}

Novel view synthesis (NVS) aims to render previously unseen viewpoints of a scene from sparse observations. Existing generalizable methods broadly follow two complementary paradigms. Reconstruction-oriented approaches infer explicit or implicit scene representations, including neural radiance fields, Gaussian primitives, and feed-forward 3D representations~\citep{mildenhall2021nerf,kerbl2023gaussians,charatan2024pixelsplat,chen2024mvsplat}, while Transformer-based models such as LVSM~\citep{jin2025lvsm}, RayZer~\citep{jiang2025rayzer}, LagerNVS~\citep{szymanowicz2026lagernvs}, and SVSM~\citep{kim2026svsm} learn view synthesis with reduced explicit 3D inductive bias. These approaches provide strong geometric fidelity when target views are well supported by observations, but sparse inputs inevitably leave parts of the scene unobserved, making deterministic reconstruction increasingly under-constrained under large viewpoint changes. In unseen regions, predictions can degenerate into patch-like artifacts biased toward colors observed in nearby source views.

A complementary line formulates NVS as conditional generation and exploits image or video diffusion priors to complete unseen content~\citep{watson2023novel,kong2024eschernet,zheng2024free3d,muller2024multidiff,yu2025viewcrafter,zhou2025seva,wu2026framecrafter}. Such models provide strong generative priors, but often lack explicit geometric conditioning to anchor generation to observed scene structure. As a result, generative priors may override geometry-supported evidence and hallucinate plausible yet inconsistent content, leading to structural drift, unstable occlusions, and cross-view inconsistency under wide-baseline interpolation and extrapolation. Meanwhile, visual geometry foundation models such as VGGT~\citep{wang2025vggt} and VGGT-$\Omega$~\citep{wang2026vggtomega} learn rich multi-view representations encoding appearance, 3D structure, correspondence, and confidence, offering a natural way to ground generative NVS with explicit geometric evidence.

These limitations suggest that generative NVS should combine strong completion priors with explicit yet uncertainty-aware geometric guidance. We introduce \method{}, a geometry-routed multi-view diffusion framework that injects visual geometry latents from VGGT-$\Omega$ into a pretrained video diffusion model. Rather than rediscovering geometry from RGB and camera rays alone, \method{} associates visual tokens with 3D locations and confidence to construct spatially aligned conditions for source and query views. Since hard projection is brittle near occlusions and uncertain geometry, we propose a confidence-aware Visual Geometry Router (VGR) that preserves front and back surface evidence together with visibility uncertainty, providing query-aligned guidance without treating reconstructed geometry as a complete scene explanation.

A lightweight input projection maps the concatenated source RGB latents,
Pl\"ucker ray maps, noisy query latents, and routed geometry conditions into
the pretrained DiT latent space, allowing \method{} to reuse the video
diffusion prior with minimal architectural modification. \method{} then jointly
denoises multiple query views within the same DiT sequence, enabling direct
information exchange across targets. Joint generation alone, however, does not
guarantee consistency with a shared 3D scene. We therefore introduce
Point-Track Residual Consistency (PTRC), which uses VGGT-derived 3D
correspondences to align denoising residual errors of the same physical point
across generated views, rather than forcing view-dependent features or VAE
latents to match. Since geometry recovered from sparse observations is
inevitably imperfect, we randomly retain, attenuate, or remove the routed
geometry condition during training to prevent over-reliance on uncertain
projections. This regularization also enables optional Geometry-Prior CFG at
inference, which strengthens geometry-aware denoising and further improves
novel-view synthesis quality. In this way, visual geometry both conditions the
diffusion process and regularizes multi-view generation.

Together, \method{} combines geometry foundation priors with video diffusion
for faithful and generative sparse-view NVS. Our contributions are threefold:
\textbf{(1) Geometry-routed generation.} A confidence-aware Visual Geometry
Router transforms VGGT-$\Omega$ features and 3D locations into query-aligned
conditions that ground generative completion in observed scene structure
without treating recovered geometry as a complete scene representation;
\textbf{(2) Geometry-grounded consistency.} Point-Track Residual Consistency
aligns predicted-clean residuals along reliable 3D correspondences, reducing
cross-view drift without suppressing valid view-dependent appearance; and
\textbf{(3) Geometry-condition regularization.} Stochastic attenuation and
dropout reduce over-reliance on imperfect projections and support optional
matched guidance at inference. Experiments demonstrate competitive or
state-of-the-art performance across pose difficulties and improved geometric
reconstructability of jointly generated views.

\section{Related Work}


\noindent\textbf{Generalizable and Feed-Forward View Synthesis.}
Novel view synthesis has progressed from scene-specific representations such as NeRF and Gaussian Splatting~\citep{mildenhall2021nerf,barron2021mipnerf,kerbl2023gaussians,chen2025slgaussian,chen2026quantifying} toward generalizable models that amortize reconstruction across scenes. Early approaches infer radiance fields or image-based representations through pixel-aligned features, multi-view stereo, epipolar reasoning, or learned ray aggregation~\citep{yu2021pixelnerf,chen2021mvsnerf,wang2021ibrnet,varma2023gnt}. More recent methods reduce explicit 3D inductive bias and learn view synthesis with Transformers, including SRT~\citep{sajjadi2022srt}, ViewFormer~\citep{kulhanek2022viewformer}, LVSM~\citep{jin2025lvsm}, Efficient-LVSM~\citep{jia2026efficientlvsm}, and RayZer~\citep{jiang2025rayzer}. In parallel, feed-forward models directly predict renderable 3D representations~\citep{charatan2024pixelsplat,chen2024mvsplat,szymanowicz2024splatter,hong2024lrm,zhang2024gslrm}, while recent works further improve geometry-aware scaling and efficiency through LagerNVS~\citep{szymanowicz2026lagernvs}, SVSM~\citep{kim2026svsm}, projective conditioning~\citep{wu2026projections}, and SHARP~\citep{mescheder2026sharp}. These approaches provide strong geometric fidelity and rendering, but deterministic reconstruction remains under-constrained when sparse observations do not cover content revealed by distant query views.

\noindent\textbf{Generative and Geometry-Guided Novel View Synthesis.}
Generative NVS uses image or video priors to complete unobserved
regions. Early diffusion methods perform pose-conditioned synthesis or model
joint multi-view distributions~\citep{watson2023novel,liu2023zero123,
shi2023zero123pp,shi2024mvdream,liu2024syncdreamer,ye2023consistent123,
kong2024eschernet,zheng2024free3d}, while later methods exploit video priors
for stronger cross-view coherence~\citep{kwak2024vivid,voleti2024sv3d,
muller2024multidiff,yu2025viewcrafter,zhou2025seva}. Stronger pixel-space
backbones improve end-to-end NVS~\citep{Elata_2025_CVPR}, and
FrameCrafter~\citep{wu2026framecrafter} adapts pretrained video diffusion to
complete unordered posed views. Related work studies hybrid deterministic and
generative modeling~\citep{le2025umami}, test-time video
completion~\citep{xu2025glimpses}, dynamic arbitrary-view
generation~\citep{vanhoorick2026anyview}, and correspondence-supervised
diffusion~\citep{Kwon_2026_CVPR}. However, source-to-query correspondence often
remains implicit, limiting robustness to large viewpoint changes and
cross-view consistency. Geometry foundation models such as
DUSt3R~\citep{wang2024dust3r}, MASt3R~\citep{leroy2024mast3r},
VGGT~\citep{wang2025vggt}, and VGGT-$\Omega$~\citep{wang2026vggtomega}
encode multi-view geometry, correspondence, and cameras.
Their geometric evidence has been combined with Gaussian reconstruction and
latent video diffusion~\citep{chen2024mvsplat360}, sparse-view 3D
optimization~\citep{wang2024diffusionpriors}, joint image and geometry
diffusion~\citep{kwak2026aligned}, latent generative
refinement~\citep{hirschorn2026splatent}, geometry-conditioned video
diffusion~\citep{kang2026geonvs}, and wide-baseline
guidance~\citep{zhou2026uniworldviewlargebaselineviewsynthesis}.
Unlike methods that construct complete Gaussian or radiance-field scenes,
\method{} routes visual geometry features, 3D points, and confidence into
query-aligned diffusion conditions and reuses their correspondences to
regularize jointly generated views. Geometry therefore guides and constrains
the generative prior as uncertainty-aware evidence rather than replacing it
with deterministic reconstruction.

\section{Method}
\label{sec:method}

\begin{figure*}[t]
    \centering
    \includegraphics[width=\textwidth]{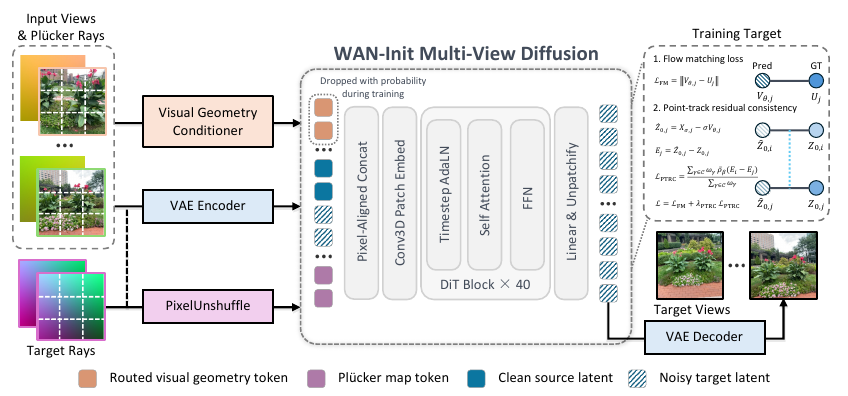}
    \caption{
    \textbf{Overview of \method{}.}
    The Visual Geometry Conditioner uses VGGT-$\Omega$ points and confidence
    to route appearance-bearing source features into query-aligned conditions.
    The diffusion state, clean-source image condition, Pl\"ucker rays, and
    routed visual condition are concatenated channel-wise, embedded
    independently within each view, and jointly processed by a Wan-initialized
    multi-view DiT. Geometry conditioning is stochastically regularized during
    training, while flow matching and PTRC supervise target fidelity and
    cross-view consistency, respectively.
    }
    \label{fig:pipeline}
    \vspace{-1.5em}
\end{figure*}

Given sparse observations, novel-view synthesis must preserve the scene
evidence visible in the source images while completing unobserved regions.
Reconstruction models encode correspondence explicitly but have limited support
for such completion; video diffusion models provide strong generative priors
but leave source-to-query correspondence largely implicit. \method{} uses
geometry to connect these two capabilities. Rather than treating the recovered
geometry as a complete scene representation, we use it to route visual evidence
into a pretrained multi-view diffusion model and to regularize the generated
views along shared 3D point tracks. The pipeline is shown in
Fig.~\ref{fig:pipeline}.

\subsection{Camera-Conditioned Multi-View Diffusion}
\label{sec:joint_diffusion}

We consider sparse-view novel-view synthesis from $M$ posed source images to
$N$ prescribed query cameras. The observed views and requested cameras are
represented as
\begin{equation}
    \mathcal{S}
    =\bigl\{(\mathbf{I}_i^{\mathrm{s}},
    \mathbf{K}_i^{\mathrm{s}},\mathbf{T}_i^{\mathrm{s}})\bigr\}_{i=1}^{M},
    \quad
    \mathcal{Q}
    =\bigl\{(\mathbf{K}_j^{\mathrm{q}},
    \mathbf{T}_j^{\mathrm{q}})\bigr\}_{j=1}^{N}.
    \label{eq:task_definition}
\end{equation}
Our goal is to synthesize query views
$\{\widehat{\mathbf{I}}_j^{\mathrm{q}}\}_{j=1}^{N}$. We place source and query views in
one diffusion sequence, enabling information exchange across slots while
preserving view-specific camera controls. A frozen VAE encodes training views
into the clean latent volume
$\mathbf{Z}_0=[\mathbf{Z}_0^{\mathrm{s}};
\mathbf{Z}_0^{\mathrm{q}}]$. 

\begin{samepage}
Following Wan's linear flow formulation~\citep{wan2025}, we sample a flow
time $\sigma$ and Gaussian noise
$\boldsymbol{\epsilon}\sim\mathcal{N}(\mathbf{0},\mathbf{I})$, and define
the noised latent and flow-matching target as
\begin{equation}
\mathbf{X}_{\sigma}=(1-\sigma)\mathbf{Z}_{0}
+\sigma\boldsymbol{\epsilon},
\quad
\mathbf{U}=\boldsymbol{\epsilon}-\mathbf{Z}_{0},
\quad
\mathcal{L}_{\mathrm{FM}}
=\frac{1}{N}\sum_{j=1}^{N}
\left\lVert\mathbf{V}_{\theta,j}-\mathbf{U}_{j}\right\rVert_{2}^{2}.
\label{eq:flow_loss}
\end{equation}
\end{samepage}

Here, $\mathbf{X}_\sigma$ is the diffusion state and
$\mathbf{V}_{\theta,j}$ is the predicted velocity for query view $j$. The
loss applies only to query slots. Query images define training targets but are
never exposed as conditions.

The model receives three complementary conditions in addition to
$\mathbf{X}_\sigma$. First, Wan's image-conditioning stream contains clean
source latents and blank query slots, providing observed appearance without
leaking query RGB. Second, dense Pl\"ucker ray maps encode the camera associated
with every source and query pixel. Third, the geometry-routed features
introduced in Sec.~\ref{sec:omega_condition} provide scene-specific
source-to-query correspondence. These signals are concatenated and mapped into
the pretrained Wan token space by an expanded input projection. The pretrained
image pathway is preserved, while newly introduced camera and geometry inputs
are initialized to have no effect at the start of fine-tuning. Source and query
tokens are then jointly processed by the Wan DiT. We keep spatial positional
encoding but remove temporal ordering from the view axis, since the inputs form
a set of camera observations rather than a video timeline. No scene-specific
text is used: training and inference share the same fixed empty textual context.
After joint denoising, query latents are decoded independently into the
requested views.

\subsection{Geometry-Routed Visual Conditioning}
\label{sec:omega_condition}

\noindent\textbf{Feature association and projection.}
Camera rays specify where a view is sampled, but not which source observation
supports each query location. We establish this correspondence using a frozen
VGGT-$\Omega$~\citep{wang2026vggtomega}, which associates each source feature
$\mathbf{f}_n$ with a 3D point $\mathbf{p}_n$ and confidence $c_n$. A
lightweight adapter maps these features into the diffusion conditioning space,
while the points and confidence determine where each feature is routed and how
strongly it contributes.

Source features retain their native image correspondence and are resampled
directly onto the diffusion grid. Query views instead require geometric
alignment. For query camera $j$, let $(\mathbf{R}_j,\mathbf{t}_j)$ denote its
extrinsics and $(f_{x,j},f_{y,j},c_{x,j},c_{y,j})$ its focal lengths and
principal point. The camera-space coordinates of $\mathbf{p}_n$ and its
projected query location are
\begin{equation}
\mathbf{x}_{nj}^{\mathrm{c}}
=(x_{nj}^{\mathrm{c}},y_{nj}^{\mathrm{c}},z_{nj}^{\mathrm{c}})^\top
=\mathbf{R}_j\mathbf{p}_n+\mathbf{t}_j,
\quad
\mathbf{u}_{nj}
=\pi_j(\mathbf{p}_n)
=\left(
f_{x,j}\frac{x_{nj}^{\mathrm{c}}}{z_{nj}^{\mathrm{c}}}+c_{x,j},
f_{y,j}\frac{y_{nj}^{\mathrm{c}}}{z_{nj}^{\mathrm{c}}}+c_{y,j}
\right)^\top.
\label{eq:projection}
\end{equation}
Here, $z_{nj}^{\mathrm{c}}$ is the query-camera depth and $\pi_j$ is perspective
projection. We route $\mathbf{f}_n$ to the resulting query-grid location,
transferring source-observed appearance evidence rather than pre-rendered RGB
values or explicit geometric predictions.

\begin{figure*}[t]
    \centering
    \includegraphics[width=0.8\textwidth]{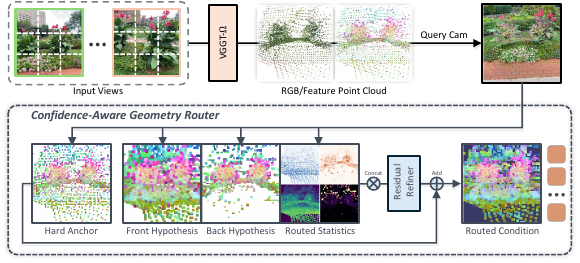}
    \caption{
    \textbf{Visual Geometry Conditioning.}
    Our conditioner constructs a query-aligned hard anchor from z-buffered
    near-front source features. A lightweight layered router forms
    confidence-weighted front and secondary hypotheses and predicts a residual
    correction from them and routing statistics.
    }
    \label{fig:geometry_conditioner}
    \vspace{-1em}
\end{figure*}

\noindent\textbf{Hard-anchor routing.}
The primary geometry condition is a depth-selected hard anchor. After
projection, each source feature is assigned to its nearest query-grid cell.
For every cell, a hard z-buffer retains features close to the nearest valid
camera-space depth and averages them according to their normalized
VGGT-$\Omega$ confidence, producing $\mathbf{G}_j^{\mathrm{hard}}$. This route
is sharp, simple, and already provides a strong source-to-query correspondence.

\noindent\textbf{Layered residual refinement.}
The hard anchor provides the primary condition, but discrete visibility may
discard valid evidence near occlusion boundaries or under small geometric
errors. We therefore construct confidence-weighted front and secondary
hypotheses only as residual cues. For a projected feature $\mathbf{f}_n$ at
continuous grid coordinate $(u_n,v_n)$ with normalized confidence
$\widetilde c_n$, its weight and aggregated feature at cell $q$, with grid
coordinate $\mathbf{q}=(q_x,q_y)$, are
\begin{equation}
\begin{aligned}
    &w_{nq}^{\ell}
    =\widetilde c_n\,k(u_n-q_x)\,k(v_n-q_y)\,
      \nu_{nq}^{\ell},\\
    \mathbf{G}_{q}^{\ell}
    &=\frac{
      \sum_{n\in\mathcal{A}_{q}^{\ell}}
      w_{nq}^{\ell}\mathbf{f}_n
    }{
      \sum_{n\in\mathcal{A}_{q}^{\ell}}
      w_{nq}^{\ell}+\varepsilon
    },
    \quad k(t)=[1-|t|]_+.
\end{aligned}
\label{eq:confidence_splat}
\end{equation}
Here, $k$ provides bilinear support and $\nu_{nq}^{\ell}$ encodes
relative-depth visibility. The front hypothesis is anchored at the nearest
splatted depth $z_q^{\mathrm{front}}$, with
$\nu_{nq}^{\mathrm{front}}
=\exp\bigl(-[z_n/z_q^{\mathrm{front}}-1]_+/\tau\bigr)$.
The secondary hypothesis retains points satisfying
$z_n>(1+\delta)z_q^{\mathrm{front}}$, anchors them at their nearest
retained depth, and applies the same attenuation. A zero-initialized residual
refiner combines both hypotheses with statistics $\mathbf{S}_j$ describing
their support, confidence, and relative depth separation:
\begin{equation}
    \mathbf{G}_j
    =\mathbf{G}_j^{\mathrm{hard}}
    +\mathcal{R}\!\left(
      \mathbf{G}_j^{\mathrm{front}},
      \mathbf{G}_j^{\mathrm{back}},
      \mathbf{S}_j
    \right).
    \label{eq:soft2_residual}
\end{equation}
Thus, the hard route remains the initial and primary condition, while layered
evidence supplies learned corrections where useful. Unsupported locations
receive no source-specific condition and are completed by the diffusion prior.

\subsection{Point-Track Residual Consistency}
\label{sec:ptrc}

The flow-matching objective in Eq.~(\ref{eq:flow_loss}) decomposes over query
views, optimizing their individual fidelity. Such per-view supervision does
not enforce geometric coherence among the remaining errors: where sparse
observations provide weak constraints, predictions of the same physical point
can drift differently across jointly generated views even when each appears
plausible and incurs a small per-view loss. A cross-view constraint is therefore
needed, but directly matching RGB values, VAE latents, or velocities would be
overly restrictive because viewpoint-dependent illumination, visibility, and
local context legitimately alter these representations. We instead couple each
prediction through its residual to its own target, enforcing consistency only
in the error that should be removed.

For a sampled flow time, the predicted clean latent and its residual in query
view $j$ are
\begin{equation}
    \widehat{\mathbf{Z}}_{0,j}
    =\mathbf{X}_{\sigma,j}-\sigma\mathbf{V}_{\theta,j},
    \quad
    \mathbf{E}_j
    =\widehat{\mathbf{Z}}_{0,j}-\mathbf{Z}_{0,j}.
    \label{eq:ptrc_residual}
\end{equation}
We reuse the VGGT-$\Omega$ point predictions to establish tracks across every
pair of query views. A point is retained only when it projects inside both
views, lies in front of both cameras, and passes a per-view z-buffer visibility
test. For a valid track $\gamma$, its projections are mapped to the nearest
latent-grid locations $\mathbf{u}_{\gamma,a}$ and
$\mathbf{u}_{\gamma,b}$. We define
$\Delta\mathbf{E}_\gamma
=\mathbf{E}_a(\mathbf{u}_{\gamma,a})-
\mathbf{E}_b(\mathbf{u}_{\gamma,b})$, and let $\bar\rho_\beta$ denote the channel
average of the scalar Smooth-L1 penalty $\rho_\beta$. With normalized
VGGT-$\Omega$ confidence $w_\gamma$, PTRC is
\begin{equation}
\begin{aligned}
\mathcal{L}_{\mathrm{PTRC}}
=\frac{
\sum_{\gamma\in\mathcal{C}}w_\gamma\,
\bar\rho_\beta(\Delta\mathbf{E}_\gamma)
}{
\sum_{\gamma\in\mathcal{C}}w_\gamma
},
\quad
\rho_\beta(x)
=\begin{cases}
x^2/(2\beta), & |x|<\beta,\\
|x|-\beta/2, & |x|\geq\beta.
\end{cases}
\end{aligned}
\label{eq:ptrc}
\end{equation}
The quadratic region provides precise alignment for small discrepancies, while
the linear region, together with confidence weighting, limits the influence of
unreliable tracks. Since PTRC aligns errors rather than predictions,
corresponding views retain valid appearance changes without drifting
independently from their respective targets.

\subsection{Training and Inference}
\label{sec:training_inference}

All camera poses are expressed in a gauge defined solely by the source cameras.
Consequently, changing the number or subset of query cameras does not alter the
conditioning coordinate frame. We freeze the VAE and VGGT-$\Omega$, and
fine-tune the diffusion transformer together with the new input and routing
modules. The number of jointly generated query views is varied during training
to support both single-view synthesis and longer view trajectories.

\noindent\textbf{Training loss.}
The full objective combines query-wise flow matching with point-track
regularization as
$\mathcal{L}=\mathcal{L}_{\mathrm{FM}}+
\lambda_{\mathrm{PTRC}}\mathcal{L}_{\mathrm{PTRC}}$, where
$\lambda_{\mathrm{PTRC}}=0.1$ in all experiments.
The two terms serve complementary roles: $\mathcal{L}_{\mathrm{FM}}$ learns
accurate denoising for each query view, while $\mathcal{L}_{\mathrm{PTRC}}$
couples prediction errors along reliable 3D point tracks.

\noindent\textbf{Geometry condition dropout.}
Geometry reconstructed from sparse observations is informative but inevitably
imperfect. During training, we therefore randomly retain, attenuate, or remove
the routed geometry condition, while leaving source appearance and camera rays
unchanged. This prevents the diffusion model from treating projected geometry
as an infallible scene reconstruction and improves its robustness when routed
evidence is sparse, uncertain, or locally missing.

\noindent\textbf{Geometry-prior CFG.}
Geometry-condition dropout also provides a reference for selectively
strengthening the geometry prior at inference. Let $\mathbf{V}_{\mathrm{G}}$ denote the
fully conditioned velocity and $\mathbf{V}_{\mathrm{ref}}$ the reference
prediction obtained by removing only query-routed geometry. During early
denoising, we apply
$\mathbf{V}_{\mathrm{GeoCFG}}=\mathbf{V}_{\mathrm{ref}}+
s_{\mathrm{g}}(\mathbf{V}_{\mathrm{G}}-\mathbf{V}_{\mathrm{ref}})$, and otherwise retain
$\mathbf{V}_{\mathrm{G}}$. The two branches share the diffusion
state, source appearance, source-aligned visual features, camera conditions,
timestep, and empty textual context. Thus, $s_{\mathrm{g}}$ strengthens query geometry
without removing the observations that define the scene.

\section{Experiments}
\label{sec:experiments}

\subsection{Experimental Setup}
\label{sec:exp_setup}

\paragraph{Data and training.}
We train on only the 1K-scene split of DL3DV-10K
(960P)~\citep{ling2024dl3dv}. Removing one unavailable scene and 19
overlapping with the 140-scene DL3DV-Benchmark leaves 980 training scenes.
\method{} initializes from Wan2.1-I2V-14B~\citep{wan2025}. We freeze the VAE
and VGGT-$\Omega$~\citep{wang2026vggtomega}, while training the Wan DiT,
expanded input projection, visual-feature adapter, and Visual Geometry Router.
Training runs for 147 epochs at $192{\times}336$, then 60 at
$480{\times}832$, using BF16 and a global batch size of eight. We adopt a
$6$-to-variable-$N$ curriculum: each sample uses six sources and jointly
denoises $N$ targets, progressing from $N\in\{1,2,4\}$ to
$N\in\{4,8,12,16\}$. The Wan DiT uses learning rates of $10^{-5}$ and
$5{\times}10^{-6}$ across the two stages; new modules use $10^{-4}$
throughout. Additional half-resolution experiments on the full DL3DV-10K yield
substantial gains from scaling data and optimization; see the Appendix.

\paragraph{Evaluation and baselines.}
We evaluate 6,188 DL3DV-Benchmark targets and test zero-shot transfer on
Mip-NeRF 360~\citep{barron2022mipnerf360}. All methods receive identical
six-view sources and target cameras, with each target generated independently;
\method{} uses 50 flow-sampling steps. Our evaluation follows
FrameCrafter~\citep{wu2026framecrafter}, with native outputs resized or
center-cropped to $480{\times}480$. We report PSNR,
SSIM~\citep{wang2004ssim}, LPIPS~\citep{zhang2018lpips}, and
DreamSim~\citep{fu2023dreamsim}. Baselines span regression-based and diffusion-based models in the tables. Additional diagnostics use 192 pose-stratified targets and assess eight-view
geometric consistency with two frozen reconstructors, VGGT-$\Omega$ and
Pi3~\citep{wang2026pi}.

\subsection{Comparison with Baselines}
\label{sec:main_results}
\label{sec:geometric_analysis}


\begin{table*}[t]
\centering
\caption{
\textbf{Quantitative comparison of 6-view NVS on DL3DV and Mip-NeRF 360.}
All methods follow the same protocol within each benchmark; \#Scenes excludes
generic foundation-model pretraining. LPIPS uses AlexNet features, and DS
denotes DreamSim. \method{} (scaled) uses the clean 9,510-scene split, with
100K effective half-resolution steps followed by 6K full-resolution adaptation
steps. Best, second-best, and third-best results use the corresponding rank
colors.
}
\label{tab:main_dl3dv}

\fontsize{7.6pt}{8.8pt}\selectfont
\setlength{\tabcolsep}{3.4pt}
\renewcommand{\arraystretch}{1.07}
\setlength{\arrayrulewidth}{0.35pt}

\begin{adjustbox}{max width=\textwidth}
\begin{tabular}{
   L{3.65cm}
   C{4.45cm}
   C{0.95cm}
   |
   C{0.98cm} C{0.98cm} C{0.98cm} C{0.98cm}
   |
   C{0.98cm} C{0.98cm} C{0.98cm} C{0.98cm}
}
\specialrule{0.9pt}{0pt}{3pt}

\multirow{2}{*}{Method}
& \multirow{2}{*}{Backbone}
& \multirow{2}{*}{\#Scenes}
& \multicolumn{4}{c|}{DL3DV}
& \multicolumn{4}{c}{Mip-NeRF 360} \\

\cmidrule(lr){4-7}
\cmidrule(lr){8-11}

& & &
PSNR $\uparrow$
& SSIM $\uparrow$
& LPIPS $\downarrow$
& DS $\downarrow$
&
PSNR $\uparrow$
& SSIM $\uparrow$
& LPIPS $\downarrow$
& DS $\downarrow$ \\

\specialrule{0.65pt}{2pt}{2pt}

\multicolumn{11}{l}{\textit{Regression-based Models}} \\

AnySplat~\citep{jiang2025anysplat}
& VGGT~\citep{wang2025vggt}
& 254K
& 12.388 & 0.298 & 0.510 & 0.214
& 11.009 & 0.232 & 0.594 & 0.252 \\

E-RayZer~\citep{zhao2026erayzer}
& --
& 10K
& 16.850 & 0.442 & 0.455 & 0.254
& \second{16.560} & \third{0.343} & 0.621 & 0.340 \\

LVSM~\citep{jin2025lvsm}
& --
& 67.5K
& 17.090 & 0.478 & 0.333 & 0.204
& 15.250 & 0.317 & 0.609 & 0.577 \\

DepthSplat~\citep{xu2025depthsplat}
& Depth Anything V2~\citep{yang2024depthanythingv2}
& 77.5K
& \third{17.284} & \second{0.566} & 0.307 & 0.158
& 15.977 & \first{0.370} & 0.410 & 0.215 \\

\specialrule{0.35pt}{2pt}{2pt}

\multicolumn{11}{l}{\textit{Diffusion-based Models}} \\

EscherNet~\citep{kong2024eschernet}
& SD1.5~\citep{rombach2022ldm}
& 10K
& 12.070 & 0.251 & 0.484 & 0.227
& 11.140 & 0.126 & 0.540 & 0.315 \\

Aether~\citep{zhu2025aether}
& CogVideoX-5B~\citep{yang2025cogvideox}
& --
& 12.660 & 0.258 & 0.469 & 0.140
& 12.600 & 0.220 & 0.651 & 0.334 \\

MVSplat360~\citep{chen2024mvsplat360}
& SVD~\citep{blattmann2023svd}
& 69.5K
& 14.150 & 0.358 & 0.513 & 0.174
& 13.859 & 0.292 & 0.636 & 0.278 \\

SEVA~\citep{zhou2025seva}
& SD2.1~\citep{rombach2022ldm}
& 80K
& 16.150 & 0.470 & 0.253 & 0.088
& 14.590 & 0.294 & 0.372 & 0.137 \\

FrameCrafter~\citep{wu2026framecrafter}
& Wan2.1-I2V-14B~\citep{wan2025}
& 1K
& 17.180 & 0.445 & \third{0.223} & \third{0.066}
& 15.640 & 0.279 & \third{0.365} & \third{0.111} \\

\method{}
& Wan2.1-I2V-14B~\citep{wan2025}
& 1K
& \second{18.104} & \third{0.508} & \second{0.222} & \second{0.062}
& \third{16.296} & 0.317 & \second{0.315} & \second{0.091} \\

\method{} (scaled)
& Wan2.1-I2V-14B~\citep{wan2025}
& 10K
& \first{19.283} & \first{0.569} & \first{0.198} & \first{0.054}
& \first{17.217} & \second{0.362} & \first{0.283} & \first{0.080} \\

\specialrule{0.9pt}{3pt}{0pt}
\end{tabular}
\end{adjustbox}

\vspace{-1.5em}
\end{table*}

\noindent\textbf{(1) Overall performance.}
Table~\ref{tab:main_dl3dv} compares all methods under the common protocol.
On 6,188 DL3DV targets, \method{} ranks first in PSNR, LPIPS, and DreamSim
and second in SSIM. Notably, these results are achieved using only $\sim$1k
training scenes, substantially fewer than many competing methods, demonstrating
strong data efficiency. Against FrameCrafter with the same Wan2.1-I2V-14B
prior and a comparable training scale, \method{} gains $0.924$ dB PSNR and
$0.063$ SSIM while retaining comparable perceptual quality, demonstrating the
benefit of geometry-routed adaptation. On zero-shot Mip-NeRF 360, \method{}
ranks first in LPIPS and DreamSim and second in PSNR. Although E-RayZer and DepthSplat lead PSNR and SSIM, respectively, their substantially higher perceptual errors indicate that these gains come at the
cost of perceptual quality.

\begin{figure*}[t]
\centering
\includegraphics[width=\textwidth]{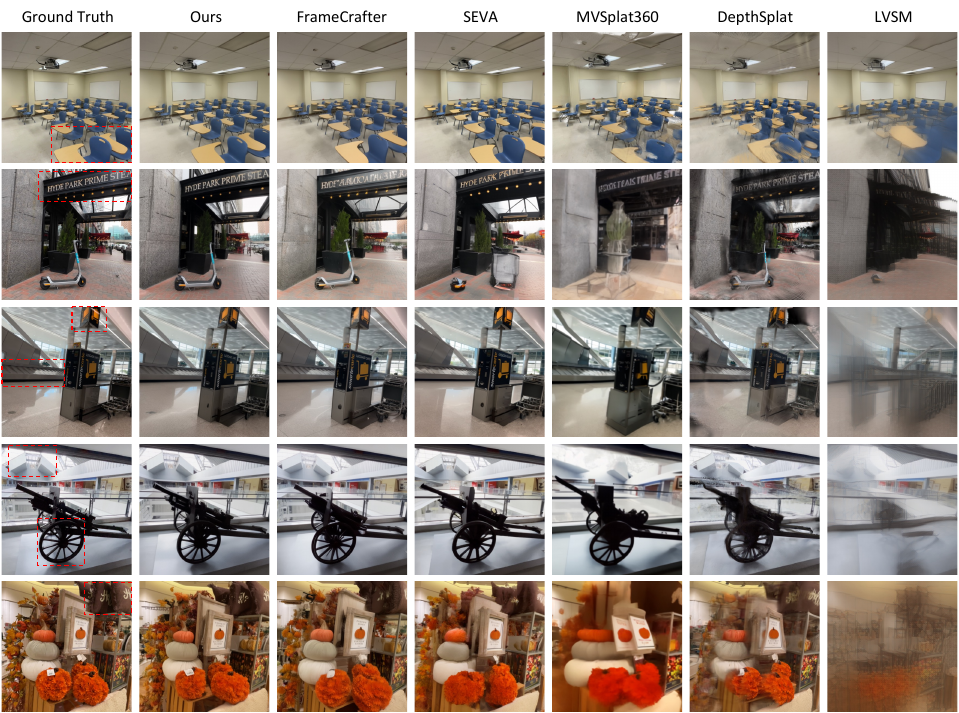}
\caption{
\textbf{Qualitative comparison on diverse DL3DV scenes.}
\method{} better matches the requested viewpoints while preserving fine
geometry, legible text, and object identity; previous methods exhibit blur,
structural distortion, or content drift.
}
\label{fig:dl3dv_qualitative}
\vspace{-1.25em}
\end{figure*}

\begin{figure*}[t]
\centering
\includegraphics[width=\textwidth]{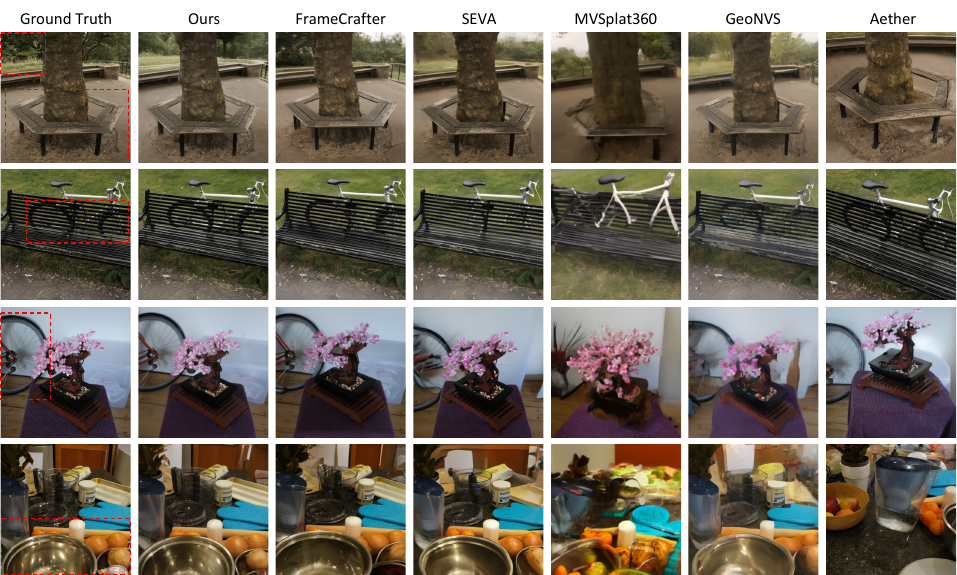}
\caption{
\textbf{Qualitative comparison on Mip-NeRF 360.}
Without dataset-specific fine-tuning, \method{} better preserves fine
structures, occlusion boundaries, and source-observed appearance while more
accurately matching the target view.
}
\label{fig:mipnerf360_qualitative}
\vspace{-1em}
\end{figure*}

\textbf{(2) Qualitative comparison.}
Fig.~\ref{fig:dl3dv_qualitative} and Fig.~\ref{fig:mipnerf360_qualitative}
localize these gains. On DL3DV, \method{} preserves chair legs, desk
boundaries, facade geometry, and legible signage while recovering target
layout and occlusion order; competing methods blur thin structures, distort
geometry, or drift from the requested camera. On zero-shot Mip-NeRF 360,
\method{} likewise better preserves object boundaries and source-observed
appearance, indicating scene-grounded rather than unsupported synthesis.


\begin{table*}[t]
\centering

\begin{minipage}[t]{0.4657\textwidth}
\vspace{0pt}
\centering
\caption{
\textbf{PSNR across target-pose difficulty on DL3DV.}
All methods use independent $6$-to-$1$ inference over 32 targets per bin.
}
\label{tab:dl3dv_pose_difficulty_psnr}

\vspace{-0.35em}
\fontsize{6.8pt}{7.8pt}\selectfont
\setlength{\tabcolsep}{1.8pt}
\renewcommand{\arraystretch}{1.04}
\setlength{\arrayrulewidth}{0.35pt}

\resizebox{\linewidth}{!}{%
\begin{tabular}{
    L{3.25cm}
    |
    C{0.72cm} C{0.72cm} C{0.72cm}
    |
    C{0.72cm} C{0.72cm} C{0.72cm}
}
\specialrule{0.9pt}{0pt}{3pt}

\multirow{2}{*}{Method}
& \multicolumn{3}{c|}{Interpolation}
& \multicolumn{3}{c}{Extrapolation} \\

\cmidrule(lr){2-4}
\cmidrule(lr){5-7}

& Near & Mid & Far
& Near & Mid & Far \\

\specialrule{0.65pt}{2pt}{2pt}

\multicolumn{7}{l}{\textit{Regression-based Models}} \\

AnySplat~\citep{jiang2025anysplat}
& 14.196 & 12.415 & 10.862
& 13.977 & 12.566 & 11.827 \\

E-RayZer~\citep{zhao2026erayzer}
& 17.388 & 14.878 & 13.824
& 16.833 & 14.477 & 14.712 \\

LVSM~\citep{jin2025lvsm}
& 20.276 & 16.599 & 13.969
& 20.790 & 16.633 & 15.579 \\

DepthSplat~\citep{xu2025depthsplat}
& 20.376 & 17.424 & \third{14.820}
& 20.653 & 17.943 & 16.374 \\

\specialrule{0.35pt}{2pt}{2pt}

\multicolumn{7}{l}{\textit{Diffusion-based Models}} \\

Aether~\citep{zhu2025aether}
& 14.030 & 12.167 & 11.065
& 15.739 & 12.295 & 11.795 \\

GEN3C~\citep{ren2025gen3c}
& 13.620 & 12.356 & 11.548
& 14.288 & 13.050 & 12.288 \\

MVSplat360~\citep{chen2024mvsplat360}
& 15.282 & 14.420 & 13.389
& 15.542 & 14.479 & 14.176 \\

FrameCrafter~\citep{wu2026framecrafter}
& 20.338 & 17.485 & 14.649
& 21.168 & 17.888 & \third{16.594} \\

SEVA~\citep{zhou2025seva}
& \second{20.940} & \third{17.861} & 14.703
& \third{22.311} & \third{18.050} & 16.575 \\

\method{}
& \third{20.920} & \second{18.391} & \second{15.853}
& \second{22.396} & \second{18.793} & \second{17.793} \\

\method{} (scaled)
& \first{22.772} & \first{19.640} & \first{16.755}
& \first{24.222} & \first{20.282} & \first{18.673} \\

\specialrule{0.9pt}{3pt}{0pt}
\end{tabular}%
}
\end{minipage}
\hfill
\begin{minipage}[t]{0.506\textwidth}
\vspace{0pt}
\centering
\caption{
\textbf{Ablations on DL3DV.}
All variants use independent $6$-to-$1$ inference on the 192
pose-stratified targets; LPIPS uses VGG features.
}
\label{tab:ablation_summary}

\vspace{-0.35em}
\fontsize{6.8pt}{7.8pt}\selectfont
\setlength{\tabcolsep}{1.8pt}
\renewcommand{\arraystretch}{1.04}
\setlength{\arrayrulewidth}{0.35pt}

\resizebox{\linewidth}{!}{%
\begin{tabular}{
    L{0.48cm}
    @{\hspace{0.08em}}
    L{2.72cm}
    |
    C{0.85cm} C{0.85cm} C{0.85cm}
}
\specialrule{0.9pt}{0pt}{3pt}

\multicolumn{2}{l|}{Configuration}
& PSNR $\uparrow$
& SSIM $\uparrow$
& LPIPS $\downarrow$ \\

\specialrule{0.65pt}{2pt}{2pt}

\multicolumn{5}{l}{\textit{Model design}} \\

w/o
& VGGT-$\Omega$ prior
& 17.294
& 0.501
& 0.357 \\

w/
& point-rendered RGB
& 18.553
& 0.536
& 0.316 \\

w/o
& PTRC
& 18.586
& 0.530
& 0.312 \\

w/o
& layered residual refinement
& 18.895
& 0.548
& 0.303 \\

\specialrule{0.35pt}{2pt}{2pt}

\multicolumn{5}{l}{\textit{Geometry-conditioning strategy}} \\

w/o
& condition regularization
& 18.722
& 0.538
& 0.309 \\

w/o
& CFG (full method)
& \second{19.024}
& \second{0.557}
& \second{0.301} \\

w/
& geometry-prior CFG
& \first{19.139}
& \first{0.563}
& \first{0.296} \\

\specialrule{0.9pt}{3pt}{0pt}
\end{tabular}%
}
\end{minipage}
\vspace{-1.5em}
\end{table*}

\noindent\textbf{(3) Performance across pose difficulties.}
To expose failures hidden by aggregate metrics, we stratify 192 DL3DV targets
by interpolation/extrapolation and near/mid/far difficulty (32 per bin).
Table~\ref{tab:dl3dv_pose_difficulty_psnr} shows that \method{} leads five bins
and trails SEVA by only $0.020$ dB in interpolation-near. Its margin over
FrameCrafter ranges from $0.582$ to $1.228$ dB and exceeds $0.9$ dB in five
bins, confirming broad effectiveness across pose difficulties.

\textbf{(4) Cross-view consistency.}
Single-target metrics cannot determine whether jointly generated views support
a coherent scene. We therefore reconstruct the same eight views with two
frozen geometry models: VGGT-$\Omega$ as the primary probe and Pi3, unused in
training or conditioning, as an independent check for shared-backbone
evaluator bias.
\textbf{\textit{If generated views are mutually consistent and match GT
content and detail, each model should recover a coherent point cloud close to
its GT-view reconstruction.}}
Across 52 DL3DV scenes, all methods use the same six sources and eight target
cameras, with generated and GT views processed identically by each
reconstructor. Table~\ref{tab:joint8_geometry} shows that \method{} ranks first
under both models. Relative to FrameCrafter, it reduces Chamfer-L1 by $26.7\%$
with VGGT-$\Omega$ and $16.4\%$ with Pi3, while improving F-score@1\% and
F-score@2\% by $11.45$/$12.48$ and $9.49$/$11.41$ percentage points,
respectively. This agreement indicates that the gains do not arise solely from
reusing VGGT-$\Omega$ as the evaluator. Appendix
Fig.~\ref{fig:joint8_pointcloud_consistency} provides corresponding
visualizations. Since the references are reconstructed from target RGB rather
than physical scans, these metrics measure relative consistency and
reconstructability, not absolute 3D accuracy.


\begin{table}[t]
\centering
\caption{
\textbf{Cross-reconstructor evaluation on DL3DV.}
Eight jointly generated views are reconstructed with frozen VGGT-$\Omega$ and
Pi3 and compared with GT-view references produced by the same reconstructor.
Pi3 is not used for training or conditioning \method{}. Metrics are
trajectory-normalized and should be compared within each reconstructor.
}
\label{tab:joint8_geometry}

\fontsize{6.8pt}{7.8pt}\selectfont
\setlength{\tabcolsep}{2.0pt}
\renewcommand{\arraystretch}{1.06}
\setlength{\arrayrulewidth}{0.35pt}

\begin{adjustbox}{max width=\columnwidth}
\begin{tabular}{
    L{3.75cm}
    |
    C{1.1cm} C{1.1cm} C{1.1cm}
    |
    C{1.1cm} C{1.1cm} C{1.1cm}
}
\specialrule{0.9pt}{0pt}{3pt}

\multirow{2}{*}{Method}
& \multicolumn{3}{c|}{VGGT-$\Omega$~\citep{wang2026vggtomega}}
& \multicolumn{3}{c}{Pi3~\citep{wang2026pi}} \\

\cmidrule(lr){2-4}
\cmidrule(lr){5-7}

& Chamfer $\downarrow$
& F@1\% $\uparrow$
& F@2\% $\uparrow$
& Chamfer $\downarrow$
& F@1\% $\uparrow$
& F@2\% $\uparrow$ \\

\specialrule{0.65pt}{2pt}{2pt}

LVSM~\citep{jin2025lvsm}
& 0.2387
& 0.1116
& 0.2218
& 0.3610
& 0.0673
& 0.1525 \\

SEVA~\citep{zhou2025seva}
& 0.1944
& 0.2050
& 0.3244
& 0.2514
& 0.1640
& 0.2751 \\

FrameCrafter~\citep{wu2026framecrafter}
& \second{0.1000}
& \second{0.3086}
& \second{0.4793}
& \second{0.1404}
& \second{0.2140}
& \second{0.3880} \\

\method{}
& \first{0.0733}
& \first{0.4231}
& \first{0.6041}
& \first{0.1174}
& \first{0.3089}
& \first{0.5021} \\

\specialrule{0.9pt}{3pt}{0pt}
\end{tabular}
\end{adjustbox}

\vspace{-1.5em}
\end{table}

\subsection{Ablation Studies}
\label{sec:ablations}

\noindent\textbf{(1) Protocol.}
All variants reported in Table~\ref{tab:ablation_summary} follow the same
full-resolution controlled protocol and use independent $6$-to-$1$ inference
on 192 pose-stratified targets. Training-time variants share the same Wan
initialization, change only the listed component, and are evaluated without
Geometry-Prior CFG. The final row applies CFG to the full checkpoint and
changes only inference. Unless stated otherwise, differences are measured
against the full checkpoint without CFG.

\textbf{(2) Model design.}
Table~\ref{tab:ablation_summary} shows that removing all
VGGT-$\Omega$-dependent components causes the largest degradation: PSNR falls
by $1.730$ dB and SSIM by $0.056$, while LPIPS rises by $0.056$. Camera rays
and the Wan prior alone therefore cannot replace visual geometry. Replacing
routed features with point-rendered RGB, while retaining the geometry, router,
and PTRC, costs $0.471$ dB PSNR and raises LPIPS by $0.015$, confirming that
appearance-bearing features provide richer evidence than incomplete RGB
renderings. Removing PTRC costs $0.438$ dB PSNR and $0.027$ SSIM, the largest
SSIM loss among single-component ablations, supporting residual alignment over
shared 3D tracks. Removing layered refinement while retaining the hard anchor
causes smaller drops of $0.129$ dB PSNR and $0.009$ SSIM, indicating that hard
routing captures most of the benefit while layered refinement corrects
ambiguous locations near occlusions and imperfect projections.

\textbf{(3) Geometry-conditioning strategy.}
Removing condition regularization costs $0.302$ dB PSNR and $0.019$ SSIM while
increasing LPIPS by $0.008$, confirming that stochastic attenuation and dropout
reduce over-reliance on imperfect geometry. The no-CFG row evaluates the full
checkpoint without guidance; applying Geometry-Prior CFG to the same checkpoint
improves PSNR by $0.115$ dB and SSIM by $0.006$ while reducing LPIPS by $0.005$.
The larger gain from regularization identifies it as the primary mechanism,
while matched CFG provides a modest complementary improvement at inference
without retraining or changing the learned model parameters or training
objective.

\textbf{Additional experiments, protocol details, results and analyses are provided in
the Appendix.}

\section{Conclusion}

We presented \method{}, a geometry-routed multi-view framework combining
visual geometry with pretrained video diffusion for sparse-view NVS. Its
confidence-aware VGR maps source features into query-aligned conditions using
3D points and confidence, while PTRC aligns predicted-clean residuals along
reliable correspondences. Stochastic training-time weakening and selective
inference-time strengthening improve robustness to imperfect geometry. Across diverse pose difficulties, \method{} achieves strong target-view fidelity and
cross-view coherence, combining geometry-grounded scene structure with the
generative prior's ability to complete unseen content.

\bibliography{iclr2027_conference}
\bibliographystyle{iclr2027_conference}

%

\clearpage
\appendix

\section{Appendix}
\suppressfloats[t]
\setcounter{topnumber}{4}
\setcounter{bottomnumber}{2}
\setcounter{totalnumber}{6}
\renewcommand{\topfraction}{0.95}
\renewcommand{\bottomfraction}{0.85}
\renewcommand{\textfraction}{0.05}
\renewcommand{\floatpagefraction}{0.85}
\setlength{\textfloatsep}{7pt plus 1.5pt minus 1pt}
\setlength{\floatsep}{6.5pt plus 1.5pt minus 1pt}
\setlength{\intextsep}{6pt plus 1.5pt minus 1pt}
\setlength{\abovecaptionskip}{4pt}
\setlength{\belowcaptionskip}{0pt}
\setlength{\abovedisplayskip}{6pt plus 1.5pt minus 1.5pt}
\setlength{\belowdisplayskip}{6pt plus 1.5pt minus 1.5pt}
\setlength{\abovedisplayshortskip}{3pt plus 1pt minus 1pt}
\setlength{\belowdisplayshortskip}{4pt plus 1pt minus 1pt}

\makeatletter
\def\subsection{\@startsection{subsection}{2}{\z@}{-1.05ex plus
-0.3ex minus -.15ex}{0.4ex plus .1ex}{\normalsize\sc\raggedright}}
\makeatother

\noindent\textbf{Metric convention.}
The cross-method comparison in Table~\ref{tab:main_dl3dv} uses
LPIPS-AlexNet to follow the evaluation protocol of prior work. Controlled
ablations use LPIPS-VGG consistently within the fixed 192-case diagnostic
protocol. We label the backbone explicitly and do not compare absolute LPIPS
values across these protocols.

\noindent\textbf{Appendix organization.}
The supplementary material is organized as follows.

\begingroup
\setlength{\parindent}{0pt}
\setlength{\parskip}{0.15em}

\textbf{Secs.~\ref{app:view_sampling_protocol}
and~\ref{app:fixed_hyperparameters}.}
We specify the view-sampling protocol, camera gauge, pose-difficulty
definition, and fixed training and conditioning hyperparameters.

\textbf{Secs.~\ref{app:data_compute_scaling},
\ref{app:joint_multitarget}, and~\ref{app:omega_feature_layer}.}
We analyze data and optimization scaling, joint multi-target generation, and
the feature depth and optimization scope of VGGT-$\Omega$~\citep{wang2026vggtomega}.

\textbf{Secs.~\ref{app:pose_difficulty}
and~\ref{app:pointcloud_visualization}.}
We report complete pose-difficulty results and evaluate cross-view consistency
using two independent point-cloud reconstructors.

\textbf{Secs.~\ref{app:ptrc_velocity}
and~\ref{app:additional_diagnostics}.}
We isolate the residual formulation of PTRC and provide additional
optimization and scene-level statistical diagnostics.

\textbf{Secs.~\ref{app:sparse_input_robustness}
and~\ref{app:vae_protocol}.}
We examine robustness to fewer source views and clarify the VAE protocols used
for quantitative evaluation and continuous-trajectory generation.

\textbf{Sec.~\ref{app:future_directions}.}
We conclude with promising directions for extending geometry-routed
multi-view generation.

\endgroup

\subsection{View Sampling and Pose-Difficulty Protocol}
\label{app:view_sampling_protocol}

\noindent\textbf{Training view sampling.}
For each training scene, we precompute 16 reproducible six-view source
bundles that rotate deterministically across epochs. Thirteen bundles use
global sampling: the ordered camera trajectory is divided into six contiguous
segments, and one frame is sampled from each segment. The remaining three use
the same stratification within a randomly selected local window. The six
selected views are shuffled before entering the model, exposing it to both
broad trajectory coverage and locally concentrated observations.

Targets are sampled online after excluding the active source frames. Local
bundles preferentially draw targets from their window and fall back to the
full scene when necessary. The target-count curriculum progresses through
$\{1,2,4\}\!\rightarrow\!\{2,4,8\}\!\rightarrow\!\{4,8,12\}
\!\rightarrow\!\{4,8,12,16\}$ in the first training stage and
$\{4,8\}\!\rightarrow\!\{4,8,12\}\!\rightarrow\!\{4,8,12,16\}$ in the
second. For eight or more targets, we always sample an ordered trajectory by
selecting evenly spaced frames within a random temporal span. Smaller target
sets use this mode with probability $0.5$ and otherwise sample views uniformly
without replacement. Training does not explicitly balance the pose-difficulty
bins used for evaluation.

\noindent\textbf{Source-defined camera gauge.}
Camera normalization depends only on the six source views. We choose the
source camera nearest their centroid as the anchor, align all cameras to its
rotation and origin, and normalize translation by the mean source-to-anchor
distance. Query cameras do not affect this transformation, so single-target
and joint inference share the same coordinate frame for a fixed source set.

\noindent\textbf{Pose-difficulty definition.}
We construct the diagnostic set from the official six-view DL3DV-Benchmark~\citep{ling2024dl3dv}
split, using its fixed source views and candidate test frames. Let
$i_{\min}$ and $i_{\max}$ be the minimum and maximum source frame indices.
A target is trajectory interpolation when
$i_{\min}<i_{\mathrm{t}}<i_{\max}$ and trajectory extrapolation when it lies
outside this interval. This definition follows the ordered capture trajectory
and does not imply containment within the 3D convex hull of the source cameras.

Difficulty is measured by the target's distance from its nearest source pose.
Let $\mathbf{c}_{\mathrm{t}},\mathbf{f}_{\mathrm{t}}$ denote the target camera
center and unit forward direction, and $\mathbf{c}_j,\mathbf{f}_j$ those of
source $j$.
We define the source-trajectory diameter as
$D_{\mathcal{S}}=\max_{j,k\in\mathcal{S}}
\lVert\mathbf{c}_j-\mathbf{c}_k\rVert_2$ and its stabilized value as
$\bar D_{\mathcal{S}}=\max(D_{\mathcal{S}},10^{-6})$. The angular difference
to source $j$ is
$\theta_j=\arccos\left[
\operatorname{clip}(\mathbf{f}_{\mathrm{t}}^\top\mathbf{f}_j,-1,1)\right]$.
Pose novelty is then
\begin{equation}
\nu_{\mathrm{t}}
=
\min_{j\in\mathcal{S}}
\sqrt{
\frac{\lVert\mathbf{c}_{\mathrm{t}}-\mathbf{c}_j\rVert_2^2}
{\bar D_{\mathcal{S}}^2}
+
\frac{\theta_j^2}{\pi^2}
}.
\label{eq:pose_novelty}
\end{equation}
This combines scene-normalized translation and viewing-direction change.
Near, mid, and far bins are defined by global pose-novelty tertiles computed
separately for interpolation and extrapolation. Their two thresholds are
$(0.1165,0.2147)$ for interpolation and $(0.0560,0.1272)$ for extrapolation.

\begin{figure}[t]
    \centering
    \includegraphics[width=\textwidth]{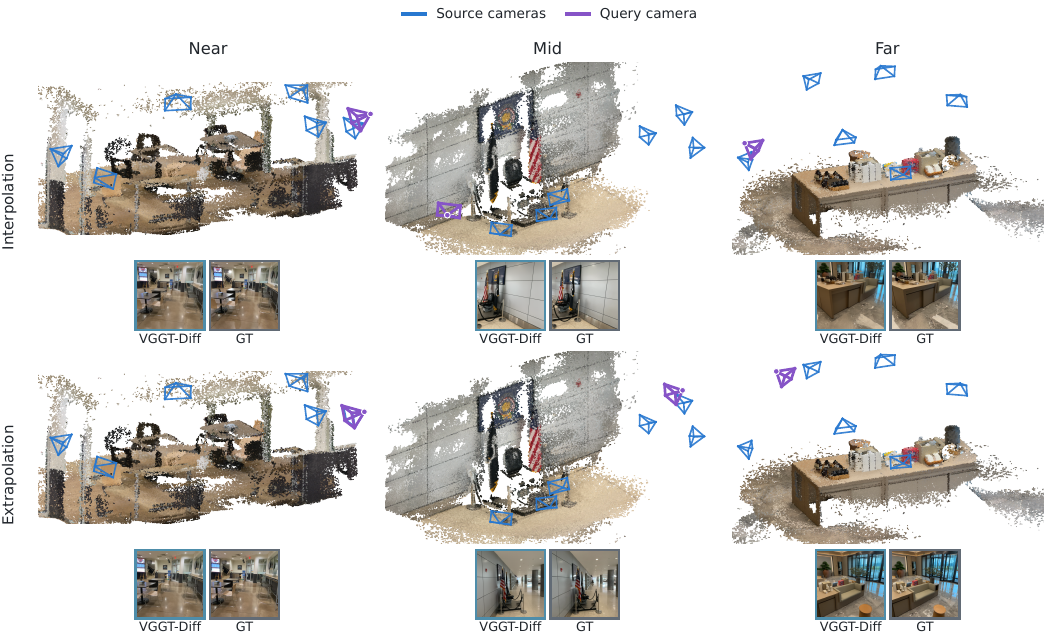}
    \caption{
    \textbf{Pose-difficulty protocol and representative predictions.}
    Blue frustums denote the six source cameras and purple denotes the query.
    Rows distinguish trajectory interpolation from extrapolation; columns
    increase pose novelty from near to far. Frozen VGGT-$\Omega$ point clouds,
    fused from multiple continuous GT-view windows, provide scene context only.
    Insets compare \method{} with GT and do not determine the bins.
    }
    \label{fig:pose_difficulty_examples}
    \vspace{-0.5em}
\end{figure}

Figure~\ref{fig:pose_difficulty_examples} visualizes the two complementary
criteria. The row depends on the target's position along the ordered capture
trajectory, whereas the column reflects its translation-and-rotation novelty
$\nu_{\mathrm{t}}$. A target can therefore be extrapolative but pose-near, or
interpolative yet pose-far.

\noindent\textbf{Frozen evaluation cases.}
The candidate pool contains 5,549 interpolation and 638 extrapolation targets.
From each of the six bins, we deterministically select 32 targets from 32
different scenes. The resulting protocol contains 192 unique targets from
109 scenes; scenes may recur across bins, but no scene-target pair is
duplicated. Every method receives the same six source views and predicts each
target independently using fixed per-case noise, without additional target
slots, rollout, or prediction feedback. This pose-stratified diagnostic is
separate from the aggregate FrameCrafter-aligned~\citep{wu2026framecrafter} comparison in
Table~\ref{tab:main_dl3dv}, and its cases remain fixed across model
resolutions.


\FloatBarrier
\subsection{Fixed Hyperparameters and Conditioning Regularization}
\label{app:fixed_hyperparameters}

Table~\ref{tab:fixed_hyperparameters} collects the constants shared by the
main experiments. The router thresholds operate on relative depth and are
therefore dimensionless; PTRC is evaluated directly in the Wan~\citep{wan2025} latent space.

\noindent\textbf{Visual Geometry Router.}
We retain the notation of Eq.~\ref{eq:confidence_splat}. The raw confidence
$c_n$ is normalized per scene as
$\widetilde c_n=\operatorname{clip}(c_n/\max_m c_m,0,1)$. For query-grid cell
$q$, the front anchor is the nearest splatted depth $z_q^{\mathrm{front}}$,
and $\nu_{nq}^{\mathrm{front}}=
\exp[-(z_n/z_q^{\mathrm{front}}-1)_+/\tau]$, with $\tau=0.02$.
The back (secondary) hypothesis retains points satisfying
$z_n>(1+\delta)z_q^{\mathrm{front}}$, where $\delta=0.04$; their nearest depth
defines $z_q^{\mathrm{back}}$, around which the same attenuation is applied.
Besides $\mathbf{G}_q^{\mathrm{front}}$ and $\mathbf{G}_q^{\mathrm{back}}$,
the router outputs four per-cell statistics: clipped front and back support,
bilinear-weighted mean confidence, and the clipped relative gap
$(z_q^{\mathrm{back}}/z_q^{\mathrm{front}}-1)$.

\noindent\textbf{PTRC optimization.}
Following Eq.~\ref{eq:ptrc_residual}, query view $j$ uses
$\widehat{\mathbf{Z}}_{0,j}=\mathbf{X}_{\sigma,j}-
\sigma\mathbf{V}_{\theta,j}$ and residual
$\mathbf{E}_j=\widehat{\mathbf{Z}}_{0,j}-\mathbf{Z}_{0,j}$.
We compute Eq.~\ref{eq:ptrc} in the Wan latent space, using the
confidence-weighted Smooth-L1 penalty with $\beta=0.1$ and no VAE decoding.
Within each training stage, $u_{\mathrm{tr}}=e/(E-1)$ denotes normalized
epoch progress and
\begin{equation}
\lambda_{\mathrm{eff}}(u_{\mathrm{tr}})
=\lambda_{\mathrm{PTRC}}
\begin{cases}
0, & u_{\mathrm{tr}}<0.1,\\
(u_{\mathrm{tr}}-0.1)/0.2,
& 0.1\leq u_{\mathrm{tr}}<0.3,\\
1, & u_{\mathrm{tr}}\geq0.3,
\end{cases}
\qquad \lambda_{\mathrm{PTRC}}=0.1.
\label{eq:ptrc_weight_schedule}
\end{equation}
Thus, PTRC is disabled for the first $10\%$ of each stage, linearly activated
during the next $20\%$, and fully weighted thereafter. The latent-space
implementation uses no additional timestep gate.

\noindent\textbf{Geometry-condition regularization.}
Each training sample draws one shared geometry-condition scale
\begin{equation}
a\sim
\begin{cases}
0, & \Pr=0.10,\\
\mathcal{U}(0.2,0.8), & \Pr=0.20,\\
1, & \Pr=0.70.
\end{cases}
\label{eq:geometry_condition_scale}
\end{equation}
The same $a$ scales the global $\Omega$ context and dense routed geometry over
both source and query slots. Source-image latents, Pl\"ucker rays, diffusion
state, and timestep remain unchanged; this operation is not text dropout.

\noindent\textbf{Geometry-Prior CFG.}
The formal guided result uses a target-only reference branch. The conditional
prediction $\mathbf{V}_{\mathrm{G}}$ retains all routed geometry, whereas
$\mathbf{V}_{\mathrm{ref}}$ zeros only query-slot dense geometry while preserving
source-aligned geometry, global $\Omega$ context, image latents, Pl\"ucker
rays, diffusion state, timestep, and empty-text context. For denoising step
$k$, with $u_{\mathrm{den}}=k/(K-1)$, we use
\begin{equation}
\mathbf{V}_{\mathrm{GeoCFG}}
=\mathbf{V}_{\mathrm{ref}}+s_{\mathrm{g}}^{\mathrm{eff}}(u_{\mathrm{den}})
(\mathbf{V}_{\mathrm{G}}-\mathbf{V}_{\mathrm{ref}}),\qquad
s_{\mathrm{g}}^{\mathrm{eff}}(u_{\mathrm{den}})
=1+(s_{\mathrm{g}}-1)\cos^2\!\left[
\frac{\pi}{2}\min\!\left(\frac{u_{\mathrm{den}}}{0.6},1\right)
\right].
\label{eq:geometry_cfg_schedule}
\end{equation}
Hence guidance decays smoothly from $2$ to $1$ over the first $60\%$ of the
50-step trajectory and is inactive thereafter; it is neither a hard switch
nor constant-scale guidance. The full-$\Omega$ CFG ablation instead removes
the global context and all source/query routed geometry from its reference
branch.

\begin{table}[!htbp]
\centering
\caption{
\textbf{Fixed geometry-conditioning hyperparameters.}
The relative-depth router constants are shared across scenes and resolutions.
}
\label{tab:fixed_hyperparameters}

\vspace{-0.3em}
\fontsize{7.2pt}{8.2pt}\selectfont
\setlength{\tabcolsep}{3.0pt}
\renewcommand{\arraystretch}{1.06}
\setlength{\arrayrulewidth}{0.35pt}

\begin{adjustbox}{max width=0.94\textwidth}
\begin{tabular}{
    L{2.35cm}
    L{4.35cm}
    C{2.65cm}
}
\specialrule{0.9pt}{0pt}{3pt}

Component
& Setting
& Value \\

\specialrule{0.65pt}{2pt}{2pt}

Visual Geometry Router
& Secondary-layer relative margin
& $\delta=0.04$ \\

Visual Geometry Router
& Relative-depth visibility temperature
& $\tau=0.02$ \\

PTRC
& Full loss weight; Smooth-L1 transition
& $0.1$; $0.1$ \\

PTRC schedule
& Stage-wise off / linear ramp / full-weight fractions
& $10\%$ / $20\%$ / $70\%$ \\

Condition regularization
& Dropped / weakened / full-condition probabilities
& $0.10$ / $0.20$ / $0.70$ \\

Geometry-Prior CFG
& Guidance scale; denoising steps; guided interval
& $2.0$; $50$; early $60\%$ \\

Text conditioning
& Prompt; text-CFG scale
& empty; $1$ \\

\specialrule{0.9pt}{3pt}{0pt}
\end{tabular}
\end{adjustbox}

\vspace{-0.8em}
\end{table}


\FloatBarrier
\subsection{Scaling with Data and Optimization Budget}
\label{app:data_compute_scaling}

\noindent\textbf{Protocol.}
We extend the full VGGT-Diff model trained on DL3DV clean-10K, containing
9,510 available scenes, from the original 18K fixed-compute budget to 100K
effective steps. Every checkpoint is evaluated on the same 192 targets, with
32 cases in each interpolation/extrapolation near, mid, and far bin. All
evaluations use six source views, $192{\times}336$ inference, a
$192{\times}192$ center crop for metric computation, seed 20260823, and 50
sampling steps. We report LPIPS-VGG throughout this diagnostic; these values
should not be compared directly with the LPIPS-AlexNet results in
Table~\ref{tab:main_dl3dv}.

The losses in Table~\ref{tab:scaling10k_checkpoints} average the ten log
entries from the 1K effective steps preceding each checkpoint. Flow loss is
the original flow-matching objective, while PTRC denotes the raw
\emph{unweighted} point-track loss before applying its training weight or
activation scale.

\begin{table}[!htbp]
\centering
\caption{
\textbf{Checkpoint scaling on DL3DV clean-10K.}
Image metrics use the fixed 192-case half-resolution protocol. Losses average
the preceding 1K effective steps; PTRC is reported before weighting.
}
\label{tab:scaling10k_checkpoints}

\vspace{-0.3em}
\fontsize{7.2pt}{8.2pt}\selectfont
\setlength{\tabcolsep}{2.4pt}
\renewcommand{\arraystretch}{1.05}
\setlength{\arrayrulewidth}{0.35pt}

\begin{adjustbox}{max width=0.88\textwidth}
\begin{tabular}{
    L{1.75cm}
    |
    C{1.15cm}
    C{1.15cm}
    C{1.15cm}
    |
    C{1.55cm}
    C{1.55cm}
}
\specialrule{0.9pt}{0pt}{3pt}

\shortstack{Effective steps}
& PSNR $\uparrow$
& SSIM $\uparrow$
& LPIPS $\downarrow$
& \shortstack{Flow loss $\downarrow$}
& \shortstack{PTRC loss $\downarrow$} \\

\specialrule{0.65pt}{2pt}{2pt}

18K
& 18.7861
& 0.5272
& 0.2793
& 0.074406
& 0.087436 \\

24K
& 19.1529
& 0.5509
& 0.2621
& 0.057234
& 0.086353 \\

30K
& 19.1980
& 0.5579
& 0.2593
& 0.056245
& 0.085368 \\

36K
& 19.6120
& 0.5770
& 0.2507
& 0.055986
& 0.084214 \\

42K
& 19.5829
& 0.5762
& 0.2541
& 0.053832
& 0.080950 \\

48K
& 19.6875
& 0.5830
& 0.2490
& 0.053431
& 0.080535 \\

54K
& 19.8138
& 0.5889
& 0.2467
& 0.053542
& 0.079932 \\

60K
& 19.4351
& 0.5725
& 0.2534
& 0.052149
& 0.077931 \\

72K
& 19.9368
& 0.5973
& 0.2411
& 0.052154
& 0.077268 \\

84K
& 20.1706
& 0.6088
& 0.2373
& 0.052251
& 0.076402 \\

88K
& 20.1350
& 0.6059
& 0.2388
& 0.052218
& 0.076123 \\

96K
& 20.2644
& 0.6146
& 0.2342
& 0.050399
& 0.074175 \\

100K
& \first{20.2780}
& \first{0.6166}
& \first{0.2335}
& \first{0.050209}
& \first{0.073912} \\

\specialrule{0.9pt}{3pt}{0pt}
\end{tabular}
\end{adjustbox}

\vspace{-0.5em}
\end{table}

\begin{figure}[!htbp]
   \centering
   \includegraphics[width=\textwidth]{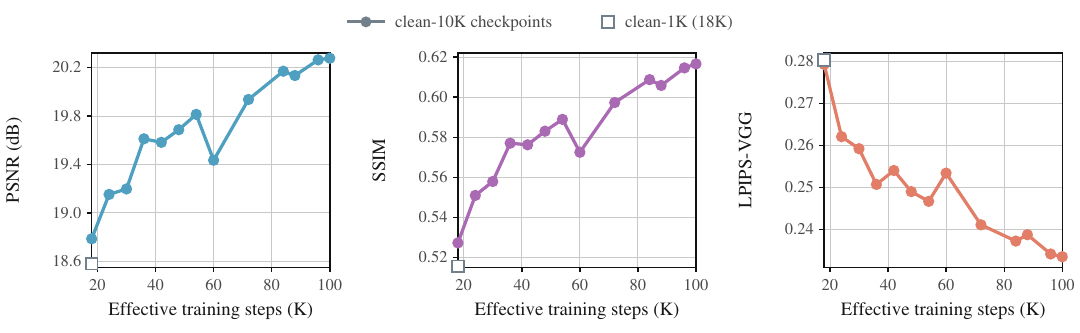}
   \caption{
    \textbf{Evaluation metrics across the clean-10K optimization budget.}
    Under the 192-case half-resolution protocol, PSNR, SSIM, and
    LPIPS-VGG are measured as training extends from 18K to 100K effective
    steps. Despite local fluctuations, all metrics improve substantially,
    with the best results at 100K. The open square denotes the clean-1K
    model evaluated at the matched 18K budget.
    }
   \label{fig:scaling10k_metrics}
   \vspace{-0.5em}
\end{figure}

\noindent\textbf{Evaluation scaling.}
Table~\ref{tab:scaling10k_checkpoints} and
Fig.~\ref{fig:scaling10k_metrics} show that the clean-10K model remains
substantially under-optimized at 18K steps. Extending training to 100K raises
PSNR from $18.7861$ to $20.2780$ dB and SSIM from $0.5272$ to $0.6166$, while
reducing LPIPS-VGG from $0.2793$ to $0.2335$. These correspond to gains of
$1.4918$ dB PSNR and $0.0894$ SSIM, together with a $0.0458$ reduction in
LPIPS-VGG.

Checkpoint quality is not strictly monotonic: 42K, 60K, and 88K exhibit local
regressions, but each is followed by recovery and the long-term trend remains
consistently favorable. The 100K checkpoint is best on all three image
metrics. Its PSNR gain over 96K narrows to $0.0135$ dB, suggesting that
optimization is approaching a plateau; continued SSIM and LPIPS-VGG
improvements prevent concluding that performance has fully saturated.

\begin{figure}[!htbp]
    \centering
    \includegraphics[width=0.94\textwidth]{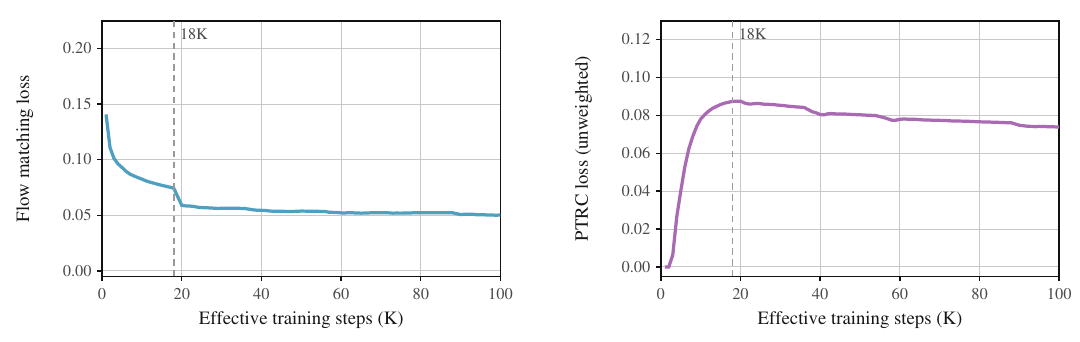}
    \caption{
    \textbf{Training dynamics on DL3DV clean-10K.}
    Curves connect non-overlapping 1K-step averages across the full training
    trajectory. The dashed line marks the original 18K fixed-compute budget.
    PTRC is reported as the raw, unweighted consistency loss. Its initial rise
    reflects scheduled activation and the increasingly difficult target
    curriculum rather than optimization divergence.
    }
    \label{fig:scaling10k_losses}
    \vspace{-1em}
\end{figure}

\noindent\textbf{Training dynamics.}
Figure~\ref{fig:scaling10k_losses} shows that optimization continues well
beyond the original fixed-compute boundary. Between the 18K and 100K
checkpoints, the preceding-1K Flow loss decreases from $0.074406$ to
$0.050209$, a $32.5\%$ reduction, while the unweighted PTRC loss decreases
from $0.087436$ to $0.073912$, a $15.5\%$ reduction. PTRC rises early because
the objective is progressively activated as the target curriculum introduces
more difficult multi-view correspondences. Once fully active, its sustained
decline accompanies the continued improvement in image quality. Together,
the evaluation and loss curves show that scaling the dataset requires a
commensurate optimization budget to realize its benefit.

\FloatBarrier

\subsection{Benefits of Joint Multi-Target Generation}
\label{app:joint_multitarget}

We compare generating eight targets in one diffusion sequence (Joint-8) with
eight strictly independent $6$-to-$1$ runs. Across 52 DL3DV scenes and 416
targets, both modes use the same checkpoint, six sources, eight target cameras,
seed, per-view noise, and 50-step sampler. Independent runs cannot access other
target cameras, latents, or predictions.

\begin{table}[!htbp]
\centering
\caption{
\textbf{Joint versus independent eight-view generation.}
Image metrics use 416 targets; geometry proxies are computed from the same
outputs with frozen VGGT-$\Omega$. LPIPS uses VGG features, and Chamfer-L1 is
trajectory-normalized.
}
\label{tab:joint_vs_independent}

\fontsize{7.2pt}{8.3pt}\selectfont
\setlength{\tabcolsep}{2.5pt}
\renewcommand{\arraystretch}{1.07}
\setlength{\arrayrulewidth}{0.35pt}

\begin{adjustbox}{max width=\textwidth}
\begin{tabular}{
L{2.75cm}
|
C{1.35cm} C{1.35cm} C{1.45cm}
|
C{1.50cm} C{1.35cm} C{1.35cm}
}
\specialrule{0.9pt}{0pt}{3pt}

Mode
& PSNR $\uparrow$
& SSIM $\uparrow$
& LPIPS $\downarrow$
& Chamfer $\downarrow$
& F@1\% $\uparrow$
& F@2\% $\uparrow$ \\

\specialrule{0.65pt}{2pt}{2pt}

Joint-8
& \first{17.6401}
& \first{0.4787}
& \first{0.3568}
& \first{0.0483}
& \first{0.3438}
& \first{0.5341} \\

Independent-8
& 17.1784
& 0.4453
& 0.3744
& 0.0741
& 0.2432
& 0.4047 \\

\specialrule{0.9pt}{3pt}{0pt}
\end{tabular}
\end{adjustbox}

\vspace{-0.25em}
\end{table}

Joint generation improves PSNR by $0.462$ dB and SSIM by $0.0334$, reduces
LPIPS-VGG by $0.0176$, lowers normalized Chamfer-L1 by $0.0259$, and raises
F-score@1\% and F-score@2\% by $0.1006$ and $0.1294$.
Scene-bootstrap 95\% confidence intervals exclude zero for all three image metrics and all three geometry metrics. Joint target slots therefore provide useful mutual
context rather than merely batching otherwise independent predictions.


\FloatBarrier
\subsection{VGGT-$\Omega$ Design Choices for Geometry Conditioning}
\label{app:omega_feature_layer}

\noindent\textbf{Setup.}
We examine feature depth and optimization scope under the same half-resolution
192-case protocol. First, we vary the extraction layer while keeping
VGGT-$\Omega$ frozen. Second, using Layer 23, we jointly optimize its visual
aggregator and encoder at a learning rate of $10^{-6}$ while retaining frozen
camera and depth heads. The Wan DiT, adapter, and Visual Geometry Router are
trained normally in all variants; all remaining inputs and objectives are
fixed.

\begin{figure}[!htbp]
    \centering
    \includegraphics[width=0.72\textwidth]{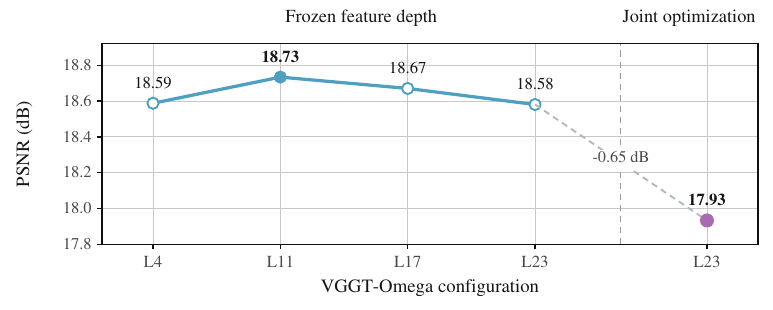}
    \caption{
    \textbf{VGGT-$\Omega$ feature depth and optimization.}
    Intermediate frozen features perform best, while jointly optimizing the
    Layer 23 visual pathway degrades PSNR.
    }
    \label{fig:omega_feature_optimization}
    \vspace{-0.5em}
\end{figure}

\begin{table}[!htbp]
\centering
\caption{
\textbf{VGGT-$\Omega$ feature and optimization diagnostics.}
All variants use the same half-resolution protocol; LPIPS uses VGG features.
}
\label{tab:omega_feature_layer}

\fontsize{7.6pt}{8.8pt}\selectfont
\setlength{\tabcolsep}{3.4pt}
\renewcommand{\arraystretch}{1.07}
\setlength{\arrayrulewidth}{0.35pt}

\begin{adjustbox}{max width=0.72\textwidth}
\begin{tabular}{
    L{4.0cm}
    C{1.35cm}
    C{1.35cm}
    C{1.45cm}
}
    \specialrule{0.9pt}{0pt}{3pt}

    Configuration
    & PSNR $\uparrow$
    & SSIM $\uparrow$
    & LPIPS $\downarrow$ \\

    \specialrule{0.65pt}{2pt}{2pt}

    \multicolumn{4}{l}{\textit{Frozen VGGT-$\Omega$: feature depth}} \\

    Layer 4
    & \third{18.588}
    & \third{0.5160}
    & 0.2828 \\

    Layer 11
    & \first{18.734}
    & \first{0.5253}
    & \first{0.2763} \\

    Layer 17
    & \second{18.670}
    & \second{0.5197}
    & \second{0.2788} \\

    Layer 23
    & 18.581
    & 0.5155
    & \third{0.2803} \\

    \specialrule{0.35pt}{2pt}{2pt}

    \multicolumn{4}{l}{\textit{Additional optimization diagnostic}} \\

    Layer 23, jointly optimized
    & 17.934
    & 0.4803
    & 0.3021 \\

    \specialrule{0.9pt}{3pt}{0pt}
\end{tabular}
\end{adjustbox}

\vspace{-0.5em}
\end{table}

\noindent\textbf{Feature depth.}
Table~\ref{tab:omega_feature_layer} shows a modest preference for intermediate
features. Layer 11 performs best across all metrics,
improving over Layer 23 by $0.153$ dB PSNR and $0.0098$ SSIM while reducing
LPIPS-VGG by $0.0040$. Intermediate representations may better balance
cross-view correspondence with local appearance: shallower
features contain less aggregated multi-view context, whereas deeper features
may trade fine detail for more abstract scene information. The limited
variation among frozen layers also indicates robustness to feature depth.

\noindent\textbf{Frozen versus jointly optimized features.}
Joint optimization reduces PSNR by $0.647$ dB and SSIM by $0.0352$, while
increasing LPIPS-VGG by $0.0218$ relative to the frozen Layer 23 setting.
Under the current clean-980 data, learning rate, and frozen-head configuration,
additional trainable capacity therefore does not improve NVS. One plausible
explanation is that flow-matching gradients alter the pretrained visual
representation while the frozen geometry heads retain their original feature
organization, producing internal drift.

These half-resolution results are not directly comparable with the
full-resolution main results. All main-paper experiments retain the
pre-specified frozen Layer 23 configuration to avoid retrospective model
selection on the fixed evaluation set. The complementary behavior across
depths motivates multi-level feature fusion as a structured alternative that
preserves the pretrained geometry representation while combining local
appearance with broader multi-view context.


\FloatBarrier
\subsection{Complete Pose-Difficulty Results}
\label{app:pose_difficulty}

We provide complete PSNR and LPIPS results across interpolation and
extrapolation difficulty. All methods use six fixed source views and
independent $6$-to-$1$ inference on a common $480{\times}480$ image plane,
with bins determined solely by camera pose. Each DL3DV bin contains 32
targets; Mip-NeRF 360 contains 66 targets per interpolation bin and 16 per
extrapolation bin. Ranking highlights follow
Table~\ref{tab:main_dl3dv}, and dashes denote unavailable results.

Unlike the aggregate comparison in Table~\ref{tab:main_dl3dv}, this
diagnostic separates trajectory relation from pose novelty. Interpolation
and extrapolation are each divided into near, mid, and far bins, revealing
whether performance degrades smoothly as target cameras move away from the
available observations. The same frozen cases are used for every method, so
differences across bins reflect pose robustness rather than sampling changes.
Figure~\ref{fig:pose_difficulty_examples} provides the corresponding camera
layouts and representative \method{} predictions for the six categories.


\begin{table}[!htbp]
\centering
\caption{
\textbf{PSNR across target-pose difficulty ($\uparrow$).}
Results are reported separately for DL3DV and Mip-NeRF 360. The scaled
checkpoint uses 100K half-resolution steps followed by 6K full-resolution
adaptation. Ties at the reported precision share the same rank.
}
\label{tab:pose_difficulty_psnr}

\fontsize{7.6pt}{8.8pt}\selectfont
\setlength{\tabcolsep}{3.4pt}
\renewcommand{\arraystretch}{1.07}
\setlength{\arrayrulewidth}{0.35pt}

\begin{adjustbox}{max width=\textwidth}
\begin{tabular}{
    L{3.65cm}
    |
    C{0.88cm} C{0.88cm} C{0.88cm}
    |
    C{0.88cm} C{0.88cm} C{0.88cm}
    |
    C{0.88cm} C{0.88cm} C{0.88cm}
    |
    C{0.88cm} C{0.88cm} C{0.88cm}
}
\specialrule{0.9pt}{0pt}{3pt}

\multirow{3}{*}{Method}
& \multicolumn{6}{c|}{DL3DV}
& \multicolumn{6}{c}{Mip-NeRF 360} \\

\cmidrule(lr){2-7}
\cmidrule(lr){8-13}

& \multicolumn{3}{c|}{Interpolation}
& \multicolumn{3}{c|}{Extrapolation}
& \multicolumn{3}{c|}{Interpolation}
& \multicolumn{3}{c}{Extrapolation} \\

\cmidrule(lr){2-4}
\cmidrule(lr){5-7}
\cmidrule(lr){8-10}
\cmidrule(lr){11-13}

& Near & Mid & Far
& Near & Mid & Far
& Near & Mid & Far
& Near & Mid & Far \\

\specialrule{0.65pt}{2pt}{2pt}

\multicolumn{13}{l}{\textit{Regression-based Models}} \\

AnySplat~\citep{jiang2025anysplat}
& 14.196 & 12.415 & 10.862
& 13.977 & 12.566 & 11.827
& 12.424 & 11.040 & 9.851
& 12.594 & 10.273 & 8.972 \\

E-RayZer~\citep{zhao2026erayzer}
& 17.388 & 14.878 & 13.824
& 16.833 & 14.477 & 14.712
& 15.964 & 15.085 & 14.028
& 15.404 & 14.760 & 12.609 \\

LVSM~\citep{jin2025lvsm}
& 20.276 & 16.599 & 13.969
& 20.790 & 16.633 & 15.579
& 16.117 & 14.488 & 13.722
& 15.296 & 14.040 & 13.016 \\

DepthSplat~\citep{xu2025depthsplat}
& 20.376 & 17.424 & \third{14.820}
& 20.653 & 17.943 & 16.374
& \third{17.478} & 16.046 & 15.303
& 16.208 & 15.518 & 12.511 \\

\specialrule{0.35pt}{2pt}{2pt}

\multicolumn{13}{l}{\textit{Diffusion-based Models}} \\

Aether~\citep{zhu2025aether}
& 14.030 & 12.167 & 11.065
& 15.739 & 12.295 & 11.795
& 12.977 & 12.994 & 12.867
& 11.841 & 10.873 & 10.894 \\

GEN3C~\citep{ren2025gen3c}
& 13.620 & 12.356 & 11.548
& 14.288 & 13.050 & 12.288
& -- & -- & --
& -- & -- & -- \\

MVSplat360~\citep{chen2024mvsplat360}
& 15.282 & 14.420 & 13.389
& 15.542 & 14.479 & 14.176
& 14.283 & 13.928 & 13.585
& 14.407 & 13.463 & 12.794 \\

FrameCrafter~\citep{wu2026framecrafter}
& 20.338 & 17.485 & 14.649
& 21.168 & 17.888 & \third{16.594}
& 16.933 & 15.700 & 15.016
& 16.222 & 15.330 & 13.017 \\

SEVA~\citep{zhou2025seva}
& \second{20.940} & \third{17.861} & 14.703
& \third{22.311} & \third{18.050} & 16.575
& \second{17.911} & \third{16.467} & \third{15.427}
& \first{17.448} & \third{15.925} & \third{13.933} \\

\method{}
& \third{20.920} & \second{18.391} & \second{15.853}
& \second{22.396} & \second{18.793} & \second{17.793}
& 17.254 & \second{16.630} & \second{15.611}
& \third{16.336} & \second{16.061} & \second{13.989} \\

\method{} (scaled)
& \first{22.772} & \first{19.640} & \first{16.755}
& \first{24.222} & \first{20.282} & \first{18.673}
& \first{18.658} & \first{17.458} & \first{16.125}
& \first{17.448} & \first{17.077} & \first{14.694} \\

\specialrule{0.9pt}{3pt}{0pt}
\end{tabular}
\end{adjustbox}

\vspace{-0.25em}
\end{table}

\noindent\textbf{Analysis.}
On DL3DV, \method{} achieves the best PSNR in five of six bins and the best
LPIPS in all six, while trailing SEVA by only $0.020$ dB in
interpolation-near. Its margin over FrameCrafter ranges from 0.582 to 1.228 dB and exceeds 0.9 dB in five of six bins. On zero-shot Mip-NeRF 360,
\method{} obtains the best LPIPS in every bin and the best PSNR in four of
six bins. The consistent perceptual gains across both datasets suggest that
the geometry-routed diffusion prior remains scene-grounded under increasing
viewpoint change.


\begin{table}[!htbp]
\centering
\caption{
\textbf{LPIPS across target-pose difficulty ($\downarrow$).}
We use VGG features on DL3DV and AlexNet features on Mip-NeRF 360. The scaled
checkpoint uses 100K half-resolution steps followed by 6K full-resolution
adaptation.
}
\label{tab:pose_difficulty_lpips}

\fontsize{7.6pt}{8.8pt}\selectfont
\setlength{\tabcolsep}{3.4pt}
\renewcommand{\arraystretch}{1.07}
\setlength{\arrayrulewidth}{0.35pt}

\begin{adjustbox}{max width=\textwidth}
\begin{tabular}{
    L{3.65cm}
    |
    C{0.88cm} C{0.88cm} C{0.88cm}
    |
    C{0.88cm} C{0.88cm} C{0.88cm}
    |
    C{0.88cm} C{0.88cm} C{0.88cm}
    |
    C{0.88cm} C{0.88cm} C{0.88cm}
}
\specialrule{0.9pt}{0pt}{3pt}

\multirow{3}{*}{Method}
& \multicolumn{6}{c|}{DL3DV}
& \multicolumn{6}{c}{Mip-NeRF 360} \\

\cmidrule(lr){2-7}
\cmidrule(lr){8-13}

& \multicolumn{3}{c|}{Interpolation}
& \multicolumn{3}{c|}{Extrapolation}
& \multicolumn{3}{c|}{Interpolation}
& \multicolumn{3}{c}{Extrapolation} \\

\cmidrule(lr){2-4}
\cmidrule(lr){5-7}
\cmidrule(lr){8-10}
\cmidrule(lr){11-13}

& Near & Mid & Far
& Near & Mid & Far
& Near & Mid & Far
& Near & Mid & Far \\

\specialrule{0.65pt}{2pt}{2pt}

\multicolumn{13}{l}{\textit{Regression-based Models}} \\

AnySplat~\citep{jiang2025anysplat}
& 0.5052 & 0.5609 & 0.6214
& 0.5057 & 0.5483 & 0.5828
& 0.5513 & 0.5854 & 0.6135
& 0.5920 & 0.6349 & 0.6830 \\

E-RayZer~\citep{zhao2026erayzer}
& 0.5496 & 0.6357 & 0.6925
& 0.5527 & 0.6433 & 0.6508
& 0.7270 & 0.7978 & 0.8284
& 0.7494 & 0.7975 & 0.7859 \\

LVSM~\citep{jin2025lvsm}
& 0.2911 & 0.4413 & 0.5656
& 0.2375 & 0.3913 & 0.4835
& 0.4980 & 0.6461 & 0.6897
& 0.5315 & 0.6589 & 0.7058 \\

DepthSplat~\citep{xu2025depthsplat}
& 0.2688 & 0.3592 & 0.4643
& 0.2600 & 0.3141 & 0.4077
& 0.3164 & 0.4058 & 0.4555
& 0.3965 & 0.4842 & 0.5703 \\

\specialrule{0.35pt}{2pt}{2pt}

\multicolumn{13}{l}{\textit{Diffusion-based Models}} \\

Aether~\citep{zhu2025aether}
& 0.4949 & 0.5891 & 0.6405
& 0.3952 & 0.5579 & 0.6098
& 0.5731 & 0.5855 & 0.5978
& 0.5965 & 0.6857 & 0.6980 \\

GEN3C~\citep{ren2025gen3c}
& 0.5194 & 0.5808 & 0.6269
& 0.4773 & 0.5578 & 0.5990
& -- & -- & --
& -- & -- & -- \\

MVSplat360~\citep{chen2024mvsplat360}
& 0.5009 & 0.5374 & 0.5952
& 0.4855 & 0.5223 & 0.5724
& 0.5820 & 0.6453 & 0.6610
& 0.6404 & 0.6683 & 0.6743 \\

FrameCrafter~\citep{wu2026framecrafter}
& 0.2489 & 0.3558 & 0.4689
& 0.2150 & 0.3011 & 0.3996
& 0.2712 & 0.3639 & 0.3996
& 0.3148 & 0.4204 & 0.5152 \\

SEVA~\citep{zhou2025seva}
& \third{0.2460} & \third{0.3390} & \third{0.4546}
& \third{0.2122} & \third{0.2959} & \third{0.3929}
& \third{0.2564} & \third{0.3267} & \third{0.3854}
& \third{0.2983} & \third{0.4013} & \third{0.4744} \\

\method{}
& \second{0.2423} & \second{0.3227} & \second{0.4163}
& \second{0.1922} & \second{0.2732} & \second{0.3572}
& \second{0.2474} & \second{0.3085} & \second{0.3504}
& \second{0.2922} & \second{0.3685} & \second{0.4424} \\

\method{} (scaled)
& \first{0.2035} & \first{0.2831} & \first{0.3861}
& \first{0.1628} & \first{0.2322} & \first{0.3248}
& \first{0.2085} & \first{0.2772} & \first{0.3236}
& \first{0.2645} & \first{0.3303} & \first{0.4122} \\

\specialrule{0.9pt}{3pt}{0pt}
\end{tabular}
\end{adjustbox}

\vspace{-0.25em}
\end{table}


\subsection{Point-Cloud Reconstruction}
\label{app:pointcloud_visualization}

\begin{figure}[!htbp]
    \centering
    \includegraphics[width=\linewidth]{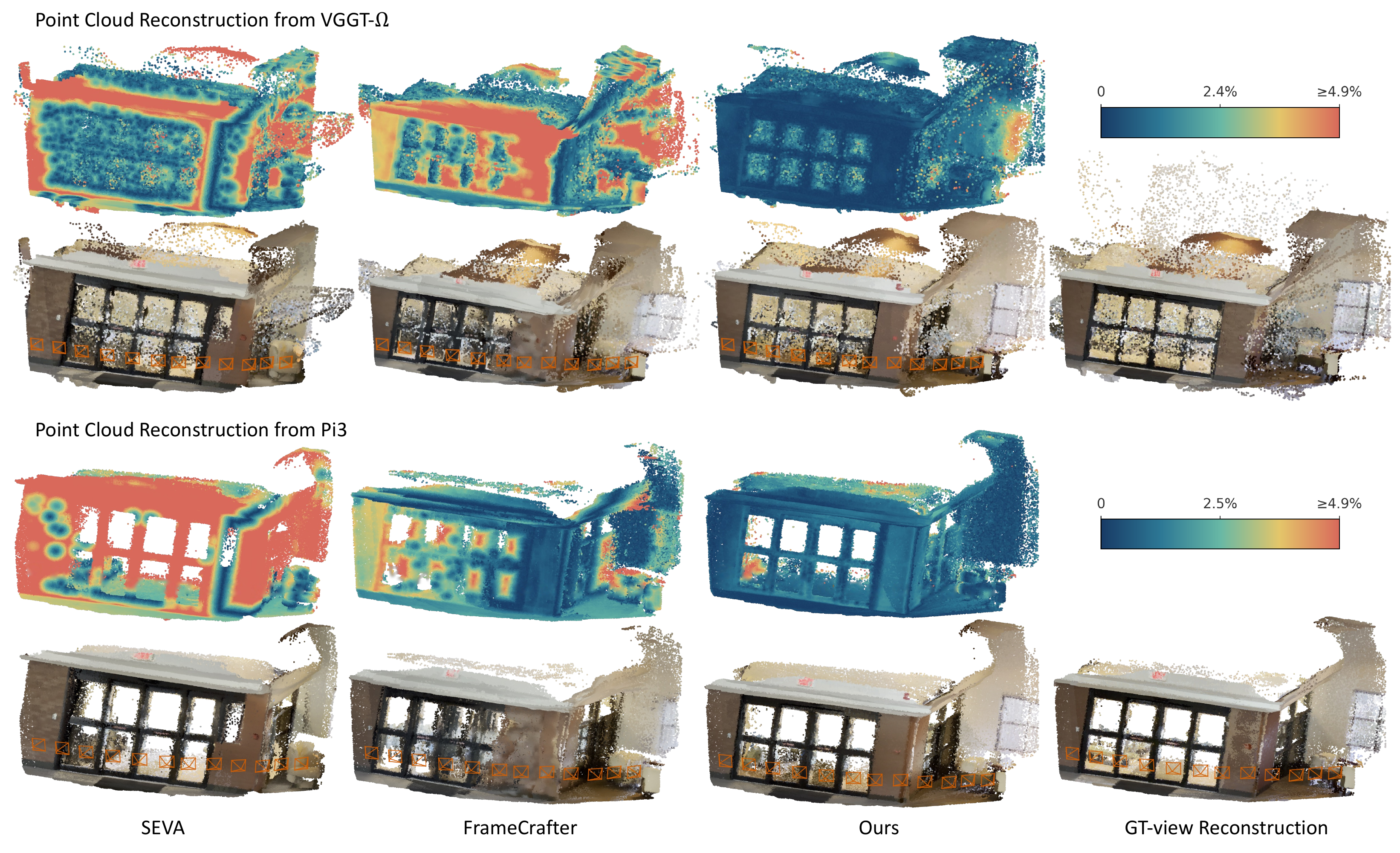}
    \caption{
    \textbf{Point-cloud reconstruction from jointly generated views.}
    The upper and lower blocks use VGGT-$\Omega$ and Pi3, respectively.
    Within each block, normalized point-to-reference errors are shown above
    the corresponding reconstructions using the displayed color scale.
    Both reconstructors show fewer displaced points and more coherent
    geometry for \method{}, indicating stronger cross-view consistency.
    }
    \label{fig:joint8_pointcloud_consistency}
    \vspace{-1em}
\end{figure}

Figure~\ref{fig:joint8_pointcloud_consistency} complements the quantitative
geometry evaluation with point clouds reconstructed from jointly generated
views. We use two frozen geometry models: VGGT-$\Omega$ as the primary probe
and Pi3~\citep{wang2026pi} as an independent reconstructor that is not used by \method{} for
training or conditioning. Their agreement helps distinguish genuine
cross-view improvements from potential shared-backbone evaluator bias.

A coherent set of generated views provides mutually compatible evidence and
should produce a stable point cloud close to the corresponding GT-view
reconstruction. Cross-view drift instead produces duplicated, displaced, or
fragmented structures. All methods use the same six source views and eight
target cameras. Within each reconstructor, generated and GT views follow
identical reconstruction settings and the same downstream alignment,
filtering, and sampling procedure. Because VGGT-$\Omega$ and Pi3 produce
different point-cloud representations, comparisons are made within each
reconstructor rather than across them. Their GT-view reconstructions serve as
model-specific references, not absolute 3D scans.

Both reconstructors produce the same overall ordering summarized in
Table~\ref{tab:joint8_geometry}. SEVA~\citep{zhou2025seva} and FrameCrafter exhibit larger
high-error regions, displaced foreground points, and less stable facade
structure. In contrast, \method{} produces more compact reconstructions with
fewer structural offsets and better preserves the shared layout across views.
The agreement between VGGT-$\Omega$ and Pi3 supports the Chamfer-L1 and F-score
gains while reducing dependence on any single geometry evaluator.

\FloatBarrier
\subsection{Residual versus Raw-Velocity Consistency}
\label{app:ptrc_velocity}

\noindent\textbf{Motivation.}
PTRC regularizes predicted-clean residuals rather than directly tying raw
model outputs across views. To isolate this design choice, we replace PTRC
with a naive objective that applies the same point tracks, confidence weights,
and robust penalty directly to predicted velocities:
\begin{equation}
\mathcal{L}_{\mathrm{Vel}}
=
\frac{
\sum_{\gamma\in\mathcal{C}} w_\gamma\,
\bar\rho_{\beta}
\left(
\mathbf{V}_{\theta,a}(\mathbf{u}_{\gamma,a})
-\mathbf{V}_{\theta,b}(\mathbf{u}_{\gamma,b})
\right)}
{\sum_{\gamma\in\mathcal{C}} w_\gamma}.
\label{eq:naive_velocity_consistency}
\end{equation}
This alternative assumes that corresponding locations should share the same
velocity. Under linear flow matching, however, each view has its own target
$\mathbf{U}_j=\boldsymbol{\epsilon}_j-\mathbf{Z}_{0,j}$; view-specific clean
latents and noise states therefore make raw velocity equality generally
invalid. Since
$\mathbf{E}_j=-\sigma(\mathbf{V}_{\theta,j}-\mathbf{U}_j)$, PTRC instead
aligns errors relative to each view's own flow target, preserving valid
differences between their velocities.

This comparison is stricter than simply removing PTRC. The direct-matching
baseline retains the same 3D correspondences and explicitly couples the same
query locations, but changes what is required to agree. It therefore tests
whether the benefit comes from generic point-track smoothing or specifically
from aligning residual errors relative to each view's supervision target.

\noindent\textbf{Controlled comparison.}
Both variants share the data, initialization, geometry conditioning,
point-track construction, and optimization, differing only in the consistency
objective. For efficiency, they follow the same half-resolution protocol and
are evaluated on the same 192 pose-stratified targets. Geometry conditioning,
condition regularization, target curriculum, and robust penalty remain
unchanged; both objectives use the stage-wise schedule in
Eq.~\ref{eq:ptrc_weight_schedule}, with full weight $0.1$.

\begin{table}[!htbp]
\centering
\caption{
\textbf{PTRC versus direct velocity matching across pose difficulty.}
Both use the same half-resolution protocol; PTRC values are highlighted and
LPIPS uses VGG features.
}
\label{tab:ptrc_velocity_consistency}

\vspace{-0.3em}
\fontsize{7.2pt}{8.3pt}\selectfont
\setlength{\tabcolsep}{2.2pt}
\renewcommand{\arraystretch}{1.06}
\setlength{\arrayrulewidth}{0.35pt}

\begin{minipage}[t]{0.487\textwidth}
\vspace{0pt}
\centering

\resizebox{\linewidth}{!}{%
\begin{tabular}{
    L{1.45cm}
    |
    C{0.95cm} C{0.95cm}
    |
    C{0.95cm} C{0.95cm}
    |
    C{0.95cm} C{0.95cm}
}
\specialrule{0.9pt}{0pt}{3pt}

\multicolumn{7}{c}{\textit{Interpolation}} \\

\specialrule{0.35pt}{2pt}{2pt}

\multirow{2}{*}{Difficulty}
& \multicolumn{2}{c|}{PSNR $\uparrow$}
& \multicolumn{2}{c|}{SSIM $\uparrow$}
& \multicolumn{2}{c}{LPIPS $\downarrow$} \\

\cmidrule(lr){2-3}
\cmidrule(lr){4-5}
\cmidrule(lr){6-7}

& Direct & PTRC
& Direct & PTRC
& Direct & PTRC \\

\specialrule{0.65pt}{2pt}{2pt}

Near
& 19.008 & \first{20.133}
& 0.531 & \first{0.602}
& 0.274 & \first{0.221} \\

Mid
& 17.520 & \first{18.103}
& 0.435 & \first{0.467}
& 0.341 & \first{0.300} \\

Far
& 15.210 & \first{15.670}
& 0.308 & \first{0.352}
& 0.440 & \first{0.397} \\

\specialrule{0.9pt}{3pt}{0pt}
\end{tabular}%
}
\end{minipage}
\hfill
\begin{minipage}[t]{0.487\textwidth}
\vspace{0pt}
\centering

\resizebox{\linewidth}{!}{%
\begin{tabular}{
    L{1.45cm}
    |
    C{0.95cm} C{0.95cm}
    |
    C{0.95cm} C{0.95cm}
    |
    C{0.95cm} C{0.95cm}
}
\specialrule{0.9pt}{0pt}{3pt}

\multicolumn{7}{c}{\textit{Extrapolation}} \\

\specialrule{0.35pt}{2pt}{2pt}

\multirow{2}{*}{Difficulty}
& \multicolumn{2}{c|}{PSNR $\uparrow$}
& \multicolumn{2}{c|}{SSIM $\uparrow$}
& \multicolumn{2}{c}{LPIPS $\downarrow$} \\

\cmidrule(lr){2-3}
\cmidrule(lr){4-5}
\cmidrule(lr){6-7}

& Direct & PTRC
& Direct & PTRC
& Direct & PTRC \\

\specialrule{0.65pt}{2pt}{2pt}

Near
& 20.021 & \first{21.326}
& 0.591 & \first{0.674}
& 0.238 & \first{0.173} \\

Mid
& 17.863 & \first{18.613}
& 0.478 & \first{0.535}
& 0.308 & \first{0.260} \\

Far
& 16.994 & \first{17.641}
& 0.415 & \first{0.463}
& 0.377 & \first{0.331} \\

\specialrule{0.9pt}{3pt}{0pt}
\end{tabular}%
}
\end{minipage}

\vspace{-0.5em}
\end{table}

\noindent\textbf{Analysis.}
Across all 192 targets, PTRC improves PSNR from $17.769$ to $18.581$ dB and
SSIM from $0.460$ to $0.516$, while reducing LPIPS from $0.330$ to $0.280$.
Table~\ref{tab:ptrc_velocity_consistency} further shows consistent gains on
every metric across all six pose bins, with PSNR improvements ranging from
$0.460$ to $1.305$ dB. Because both variants use identical correspondences
and weighting, these gains isolate the residual-relative formulation rather
than the point tracks themselves. Direct velocity matching penalizes valid
differences between view-specific flow targets, whereas PTRC couples only the
prediction errors that should be removed. The improvement appears in both
interpolation and extrapolation, showing that the residual formulation is not
tied to one camera relation or difficulty range.


\FloatBarrier
\subsection{Additional Optimization and Statistical Diagnostics}
\label{app:additional_diagnostics}

\noindent\textbf{Wan adaptation.}
We compare full DiT fine-tuning with a rank-32 LoRA adaptation while retaining
the same geometry conditioner, data, initialization, objectives, and
half-resolution 192-case evaluation. Full fine-tuning improves PSNR by
$1.324$ dB and SSIM by $0.0478$, while reducing LPIPS-VGG by $0.0751$.
The large gap indicates that sparse low-rank updates are insufficient to
integrate routed geometry into the pretrained video prior under this setup.

\begin{table}[!htbp]
\centering
\caption{
\textbf{Effect of Wan adaptation strategy.}
Both variants use the same half-resolution controlled protocol; LPIPS uses
VGG features.
}
\label{tab:wan_adaptation}

\fontsize{7.6pt}{8.8pt}\selectfont
\setlength{\tabcolsep}{3.4pt}
\renewcommand{\arraystretch}{1.07}
\setlength{\arrayrulewidth}{0.35pt}

\begin{adjustbox}{max width=0.58\textwidth}
\begin{tabular}{L{3.5cm} C{1.35cm} C{1.35cm} C{1.55cm}}
\specialrule{0.9pt}{0pt}{3pt}

Wan adaptation
& PSNR $\uparrow$
& SSIM $\uparrow$
& LPIPS $\downarrow$ \\

\specialrule{0.65pt}{2pt}{2pt}

LoRA, rank 32
& 17.2563
& 0.4677
& 0.3554 \\

Full fine-tuning
& \first{18.5807}
& \first{0.5155}
& \first{0.2803} \\

\specialrule{0.9pt}{3pt}{0pt}
\end{tabular}
\end{adjustbox}

\vspace{-0.5em}
\end{table}

\noindent\textbf{Paired scene-level statistics.}
We additionally compare locally generated \method{} and FrameCrafter outputs
under the same 140-scene, 6,188-target protocol. The scene-mean PSNR gain is
$0.7893$ dB, with a 92.14\% scene win rate and a scene-bootstrap 95\%
confidence interval of $[0.7003,0.8769]$ dB. The corresponding scene-mean
changes are $+0.0330$ SSIM, $-0.0365$ LPIPS-AlexNet, and $-0.0118$ DreamSim.
These paired results show that the aggregate improvement is distributed across
scenes rather than driven by a small subset. They are reported separately from
Table~\ref{tab:main_dl3dv}, whose baseline rows follow the published
cross-method protocol.

\subsection{Robustness to Sparse Input Views}
\label{app:sparse_input_robustness}

\noindent\textbf{Protocol.}
We evaluate whether the frozen full-resolution checkpoint trained with six
source views transfers to reduced source coverage without retraining or
test-time optimization. Both \method{} and
FrameCrafter use independent $M$-to-$1$ inference
on the same 192 pose-stratified DL3DV targets, with $M\in\{3,4\}$. Inference
runs at $480{\times}832$ for 50 sampling steps, and metrics are computed on the
center $480{\times}480$ crop. Both methods use identical source images, target
cameras, and per-case noise seeds. For each target, we select $M$ views from
the six sources fixed by the main protocol using camera poses only. A source
$s$ is ranked by
$\sqrt{(\lVert\mathbf{c}_{\mathrm{t}}-\mathbf{c}_s\rVert_2/D)^2+
(\theta_{\mathrm{t}s}/\pi)^2}$, where $D$ is the maximum pairwise distance
among the six source cameras and $\theta_{\mathrm{t}s}$ is their viewing
direction difference. The selected views are restored to their registered
trajectory order, producing nested source subsets independent of image
content, predictions, and method identity.

\begin{table}[!htbp]
\centering
\caption{
\textbf{Robustness to sparse input views on DL3DV.}
Both methods use full-resolution independent $M$-to-$1$ inference on the same
192 targets. Source subsets are selected identically using only camera poses;
LPIPS uses VGG features.
}
\label{tab:sparse_input_robustness}

\vspace{-0.3em}
\fontsize{7.4pt}{8.5pt}\selectfont
\setlength{\tabcolsep}{3.0pt}
\renewcommand{\arraystretch}{1.06}
\setlength{\arrayrulewidth}{0.35pt}

\begin{adjustbox}{max width=0.82\textwidth}
\begin{tabular}{
    C{1.15cm}
    L{3.55cm}
    |
    C{1.25cm}
    C{1.25cm}
    C{1.35cm}
    C{1.45cm}
}
\specialrule{0.9pt}{0pt}{3pt}

Sources
& Method
& PSNR $\uparrow$
& SSIM $\uparrow$
& LPIPS $\downarrow$
& DreamSim $\downarrow$ \\

\specialrule{0.65pt}{2pt}{2pt}

\multirow{2}{*}{3}
& FrameCrafter~\citep{wu2026framecrafter}
& 16.9559
& 0.4647
& 0.3709
& 0.0807 \\

& \method{}
& \first{17.7372}
& \first{0.4966}
& \first{0.3391}
& \first{0.0680} \\

\specialrule{0.35pt}{2pt}{2pt}

\multirow{2}{*}{4}
& FrameCrafter~\citep{wu2026framecrafter}
& 17.8545
& 0.5076
& 0.3390
& 0.0685 \\

& \method{}
& \first{18.4380}
& \first{0.5307}
& \first{0.3135}
& \first{0.0592} \\

\specialrule{0.9pt}{3pt}{0pt}
\end{tabular}
\end{adjustbox}

\vspace{-0.6em}
\end{table}

\noindent\textbf{Results.}
Despite being trained with six sources, \method{} remains stronger under both
reduced-view settings. With three inputs, it improves over FrameCrafter by
$0.781$ dB PSNR and $0.0319$ SSIM while reducing LPIPS-VGG by $0.0318$ and
DreamSim by $0.0127$. With four inputs, the corresponding gains are $0.584$ dB,
$0.0231$, $0.0255$, and $0.0093$. The consistent improvement across all four
metrics shows that geometry-routed conditioning remains effective when the
available observations are reduced to three or four views, without adapting
the six-view-trained checkpoint.

\begin{figure}[!htbp]
    \centering
    \includegraphics[width=\linewidth]{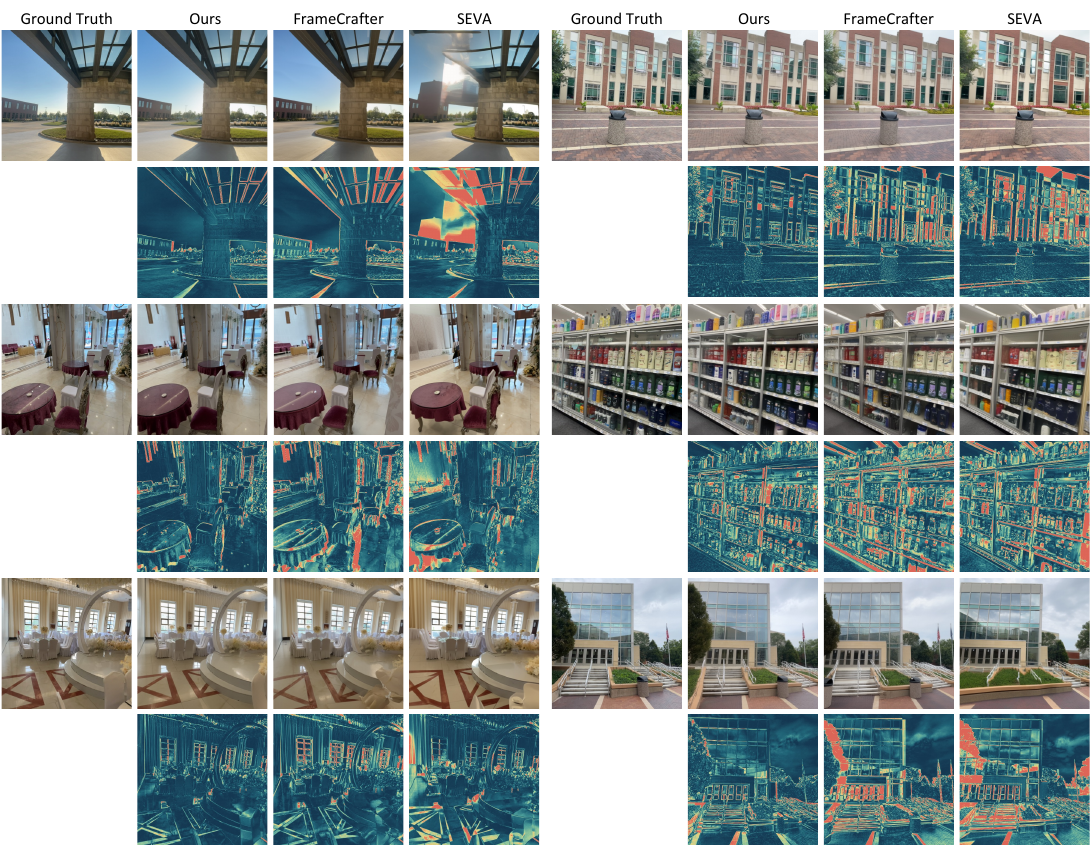}
    \caption{
    \textbf{Target-view alignment and pixel error.}
    Predicted RGB views are shown above their absolute pixel differences from
    GT under a shared error scale. \method{} produces lower and more localized
    errors than the baselines, indicating closer alignment with the GT
    viewpoint and fewer spatial or structural mismatches.
    }
    \label{fig:target_view_pixel_error}
    \vspace{-0.5em}
\end{figure}

\subsection{VAE Encoding Protocols}
\label{app:vae_protocol}

All \method{} variants use the same frozen Wan2.1 Video VAE, which produces
16-channel latents with an $8{\times}$ spatial downsampling factor
($192{\times}336 \rightarrow 16{\times}24{\times}42$ and
$480{\times}832 \rightarrow 16{\times}60{\times}104$). All quantitative
experiments adopt an independent-view latent layout. During training, each
source image and ground-truth target image is independently encoded, with the
latter providing the clean latent for denoising supervision. During inference,
only source images are encoded; target latent slots are initialized from noise,
and no target RGB image is provided to the model. Consequently, six sources
and $N$ targets produce $6+N$ view-time latent slots, each aligned one-to-one
with a physical view, camera, Pl\"ucker map, and routed geometry condition.
Thus, none of our reported quantitative results uses temporal VAE compression.

To accelerate experimentation and reduce the training cost of long sequences,
we use a hybrid protocol for the continuous 80-frame trajectory demo and
fine-tune it at the half resolution of $192{\times}336$. The six sources remain
independently encoded, while the ordered target sequence is jointly encoded
through the causal temporal pathway of the same frozen VAE, yielding
$L_T=1+\left\lceil(T-1)/4\right\rceil$ target slots. For $T=80$, we repeat the
final frame once to form an 81-frame sequence, obtaining $L_T=21$ target slots;
the repeated output is discarded after decoding. This reduces the number of
view-time latent slots processed by the DiT from $6+80=86$ to $6+21=27$.
Camera conditions bypass the VAE. Each six-channel Pl\"ucker map is spatially
rearranged into $384$ channels at latent resolution using an $8{\times}$
PixelUnshuffle. To avoid temporally discarding intermediate target poses, we
group these maps according to the causal VAE windows:
$[0]$, $[1,2,3,4]$, $\ldots$, and $[77,78,79,79]$. Within each four-frame
window, the ordered $384$-channel maps are concatenated into $1536$ channels
and projected back to $384$ by a lightweight $1{\times}1{\times}1$ Pl\"ucker
adapter, making all four poses available to the conditioning pathway while
preserving the DiT interface. The first target slot retains its single-frame
conditioning path. We initialize the adapter to select only the fourth map,
making the augmented model exactly equivalent to the previous endpoint-only
checkpoint at initialization and enabling checkpoint-compatible fine-tuning
without modifying the DiT backbone. In the current adaptation, target camera
matrices and routed geometry conditions remain aligned with the endpoint
indices $\mathcal{A}_T=\{0,4,8,\ldots,76,79\}$; only the Pl\"ucker pathway
aggregates all poses within each temporal window. Because independently
encoded and temporally compressed target latents have different temporal
semantics, we fine-tune the initialized model for this 80-frame regime rather
than treating the two representations as interchangeable. This
half-resolution extension is used only for the continuous demo and does not
affect any reported quantitative result.

\subsection{Future Directions}
\label{app:future_directions}

Our results suggest several directions for extending geometry-routed
multi-view generation. First, the complementary behavior observed across
VGGT-$\Omega$ feature depths motivates adaptive multi-level fusion that
combines fine appearance cues with increasingly global geometric context.
Second, although directly optimizing the VGGT-$\Omega$ visual pathway is
ineffective under our current setting, parameter-efficient adaptation,
geometry-preserving objectives, or staged optimization may better specialize
the encoder without disrupting its pretrained representations. The Visual
Geometry Router could also be extended beyond its hard anchor and layered
residual refinement to model richer visibility distributions, multiple
occlusion layers, and better-calibrated geometric uncertainty. At the
generation level, memory-efficient joint denoising and chunked generation may
scale the benefits of shared multi-view context to more views and longer camera
trajectories while maintaining consistency across chunks. Finally, extending
3D point tracks to spatiotemporal correspondences could generalize the
framework from static scenes to dynamic and non-rigid content with moving
objects and time-varying occlusions.

\end{document}